\documentclass[conf]{new-aiaa}
\usepackage[utf8]{inputenc}

\usepackage{graphicx}
\usepackage{amsmath}
\usepackage[version=4]{mhchem}
\usepackage{siunitx}
\usepackage{longtable,tabularx}
\usepackage{float} 
\usepackage{subfigure}
\usepackage{subfigmat}
\usepackage{booktabs}
\usepackage{caption}
\usepackage{tikz}
\usepackage{listings}
\usepackage{fancyhdr}
\usepackage{color}
\usepackage{algorithm}
\usepackage{algpseudocode}
\usepackage{bm}
\usepackage{hyperref}
\hypersetup{
  colorlinks   = true, 
  urlcolor     = blue, 
  linkcolor    = blue, 
  citecolor   = blue 
}

\definecolor{codegreen}{rgb}{0,0.6,0}
\definecolor{codegray}{rgb}{0.5,0.5,0.5}
\definecolor{codepurple}{rgb}{0.58,0,0.82}
\definecolor{backcolour}{rgb}{0.95,0.95,0.92}

\lstdefinestyle{mystyle}{
    backgroundcolor=\color{backcolour},   
    commentstyle=\color{codegreen},
    keywordstyle=\color{magenta},
    numberstyle=\tiny\color{codegray},
    stringstyle=\color{codepurple},
    basicstyle=\footnotesize,
    breakatwhitespace=false,         
    breaklines=true,                 
    captionpos=b,                    
    keepspaces=true,                 
    numbers=left,                    
    numbersep=5pt,                  
    showspaces=false,                
    showstringspaces=false,
    showtabs=false,                  
    tabsize=2
}

\title{Open-Source High-Speed Flight Surrogate Modeling Framework}

\author{Tyler E. Korenyi-Both\footnote{Phantom Fellow, Department of the Air Force-Massachusetts Institute of Technology (DAF-MIT) Artificial Intelligence (AI) Accelerator}, Nathan J. Falkiewicz\footnote{Technical Staff, Massachusetts Institute of Technology Lincoln Laboratory (MIT LL)}, Matthew C. Jones\footnote{Technical Staff, Massachusetts Institute of Technology Lincoln Laboratory (MIT LL)}}

\begin{document}

\maketitle

\begin{abstract}
High-speed flight vehicles, which travel much faster than the speed of sound, are crucial for national defense and space exploration. However, accurately predicting their behavior under numerous, varied flight conditions is a challenge and often prohibitively expensive. The proposed approach involves creating smarter, more efficient machine learning models (also known as surrogate models or meta models) that can fuse data generated from a variety of fidelity levels--to include engineering methods, simulation, wind tunnel, and flight test data--to make more accurate predictions. These models are able to move the bulk of the computation from high performance computing (HPC) to single user machines (laptop, desktop, etc.). The project builds upon previous work but introduces code improvements and an informed perspective on the direction of the field. The new surrogate modeling framework is now modular and, by design, broadly applicable to data of any shape and type, given sufficient correlation between the input and output data and adherence to the data standard presented in this work. The new framework also has a more robust automatic hyperparameter tuning capability and abstracts away most of the pre- and post-processing tasks. Models were trained, tested, and validated using unclassified flight vehicle datasets provided by the author's sponsor unit. The Gaussian process regression and deep neural network-based models included in the presented framework were able to model two datasets with high accuracy ($R^2>0.99$). The primary conclusion is that the framework is effective and has been delivered to the Air Force for integration into real-world projects. For future work, significant and immediate investment in continued research is crucial. The author recommends further testing and refining modeling methods that explicitly incorporate physical laws and are robust enough to handle simulation and test data from varying resolutions and sources, including coarse meshes, fine meshes, unstructured meshes, and limited experimental test points.  
\end{abstract}

\section*{Phantom Fellowship}
\textit{The Department of the Air Force-Massachusetts Institute of Technology Artificial Intelligence Accelerator (DAF-MIT AI Accelerator) Phantom Fellowship is a rigorous five-month career-enhancing opportunity for high-performing Department of Defense professionals to assess, develop, and gain exposure to Artificial Intelligence (AI) and Machine Learning (ML) technologies. During the program, enlisted, officer, and government civilian personnel (Phantoms) embed directly with the DAF-MIT AI Accelerator, a cutting-edge organization with the mission of advancing AI/ML technologies for the DoD while also addressing broader societal and ethical issues related to AI/ML.  During their Fellowship, Phantoms are required to apply AI/ML foundational knowledge to produce a short-form research paper that seeks to address an AI/ML problem related to their home unit and/or career field. The views expressed in this article are those of the author(s) and do not reflect the official policy or position of the United States Air Force, Department of Defense, or the U.S. Government. The inclusion of external links and references does not imply any endorsement by the author(s), the publishing unit, the Department of the Air Force, the Department of Defense or any other department or agency of the U.S. Government. They are meant to provide an additional perspective or as a supplementary resource.}

\newpage
\section{Introduction}
\lettrine{S}{ustained} atmospheric high-speed flight is a national defense priority~\cite{austin_2022,sayler_2022}. However scientific and technical organizations studying the high-speed flight regime are challenged by the harsh operating conditions~\cite{schmisseur2015hypersonics} and steep development costs~\cite{capaccio_2021, fy2022programacquisitioncostsbyweaponsystem_2021}. Limited access to wind tunnels and flight test facilities slows high speed flight research in the United States~\cite{GAO__hypersonics_2021} and increases reliance on computational analysis. Moreover, early flight vehicle design and analysis necessitate parameter sweeps over broad ranges, which is infeasible with expensive exploration techniques like high-fidelity simulations or wind tunnel testing due to costs and the inability of wind tunnel testing to fully replicate all atmospheric hypersonic flight physics over extended periods~\cite{lu_2002}. To match the data quality of wind tunnel testing and flight tests, maximizing computational model fidelity is crucial to reduce uncertainty. However, achieving high computational model fidelity often impose steep costs and resource requirements. To mitigate these challenges, classical engineering methods, such as Newtonian or panel methods, are often employed to provide quick and cost-effective solutions, though their applicability is limited and often uncertain. 

Data-driven model reduction offers a promising avenue to harness the precision of high-fidelity techniques for applications demanding rapid online predictions. Consequently, the development and understanding of data-driven methods have garnered significant attention for predicting flow environments~\cite{hall2000proper,venturi2004gappy, lucia2005aeroelastic, skujins2014reduced, rowley2017model, kutz2017deep, duraisamy2019turbulence, brunton2020machine, Dreyer2021, barnett2023neural, needels2023efficient, shukla_deep_2024}. Surrogate modeling~\cite{forrester2008engineering} is a type of data-driven model reduction that presents an opportunity to accelerate design and analysis timelines, reduce costs, and improve the quality of engineering analysis. This approach seeks to emulate the response of a system with evaluation times much faster than the high-fidelity data generation methods they are trained on and often without explicitly using the governing equations. Multi-fidelity surrogate modeling~\cite{forrester2007multi,raissi2017inferring,meng2020composite} seeks to leverage multiple training data sets of varying fidelity to achieve high-performing models without excessive high-fidelity data requirements.

Building on these concepts, physics-informed machine learning~\cite{karniadakis2021physics} incorporates information from governing equations or known physical properties, such as frame invariance or symmetry. This approach has attracted considerable attention for addressing problems typically handled with data-driven surrogate modeling~\cite{pestourie_physics-enhanced_2023, shukla_deep_2024}. Among the earliest methods, physics-informed neural networks (PINNs)\cite{raissi2019physics} explicitly encode the system's governing equations into the neural network loss function, ensuring that the model's predictions conform to physical laws. However, applying PINNs to high-dimensional and complex inverse problems presents significant challenges\cite{krishnapriyan2021characterizing}. 

Another significant advancement in this field is the use of neural operators~\cite{lu_deeponet_2021,kovachki2023neural}applied to scientific machine learning problems~\cite{azizzadenesheli2024neural}. Neural operators are machine learning models that approximate mappings between function spaces, facilitating the prediction of complex systems governed by differential equations across varying resolutions. The foundational theory underpinning neural operators~\cite{chen1995universal} was further developed and applied by Lu et al. in their seminal DeepONet paper~\cite{lu_deeponet_2021}.  As explained in the \href{https://neuraloperator.github.io/neuraloperator/dev/user_guide/neural_operators.html#operator-learning}{neuraloperator} GitHub documentation~\cite{li2020fourier, kovachki2023neural}, standard artificial neural networks map tensors to tensors (finite dimensional), similar to a .jpeg or .png, where resolution remains fixed upon zooming. In contrast, neural operators map function spaces to function spaces (infinite dimensional), analogous to a .pdf or .svg, where resolution dynamically adjusts upon zooming. Variations upon this idea include Fourier neural operators~\cite{li2020fourier,lu2022comprehensive} and physics-informed neural operators~\cite{li2021physics, goswami2023physics}. Neural operators as surrogates have been applied to high-speed flow problems, like in ~\cite{di2023neural, shukla_deep_2024}. 

Many machine learning~\cite{pedregosa2011scikit,abadi2016tensorflow,chollet2015keras} and surrogate modeling packages~\cite{lophaven2002matlab,SMT2019, saves2024smt} are available. However, the Department of the Air Force would benefit from a modular, robust, open-source, comprehensive, and most importantly simple solution. While the aforementioned machine learning and surrogate modeling packages are highly effective (and in fact most are used in the presented work), it is challenging for engineers with limited modeling experience to use these tools practically and quickly. Moreover, implementing surrogate modeling software in secure environments presents unique challenges, which an open source framework can address, though it introduces issues like complex package dependencies and the need to configure identical virtual environments on different machines. Additionally, automating pre- and post-processing tasks, which are often the most cumbersome and least documented, is crucial. To the best of the author's knowledge, no publicly available, open-source surrogate modeling packages exist with built-in features for data structuring, cross-validation, data scaling, and organized data management. 

Thus, the problem: aerospace engineers within the Department of the Air Force, and more broadly, scientific and technical organizations across the US government, need a free, open-source, and portable end-to-end multi-fidelity surrogate modeling framework.  It must be free because these organizations are often resource-constrained. It needs to be open-source, as closed-source software has lower likelihood of adoption in research organizations; if engineers cannot see or modify the source code, they're less likely to use it. Portability is essential to transfer the framework to various systems, including air-gapped environments. An end-to-end solution is necessary because the end users are often not surrogate modeling experts, making it useful to abstract unnecessary details.

The objective of this study is to enhance the predictive capabilities and reduce the computational expense of aerodynamic models for high-speed flight vehicles. To achieve this, the present study focuses on the following areas: improving the multi-fidelity modeling framework presented in~\cite{korenyi2024josr}, thus developing a robust and open-source engineering tool for use by the DoD, the intelligence community (IC), and research labs, and recommending future study of physics-based or physics-informed machine learning techniques.

\section{Methodology}

Surrogate modeling details can be found in the author's previous work \cite{korenyi2023thesis,korenyi2023conference,korenyi2024josr} and two of the seminal works in the field~\cite{Sack1989DoE, forrester2008engineering}. Details about the surrogate modeling done in this work can be found in App.~\ref{app:surrogate_modeling} and \ref{app:sm_pipeline}. 

\subsection{Data}\label{subsec:dataset}
\begin{figure}[h!]
    \centering
    \includegraphics[width=0.9\textwidth]{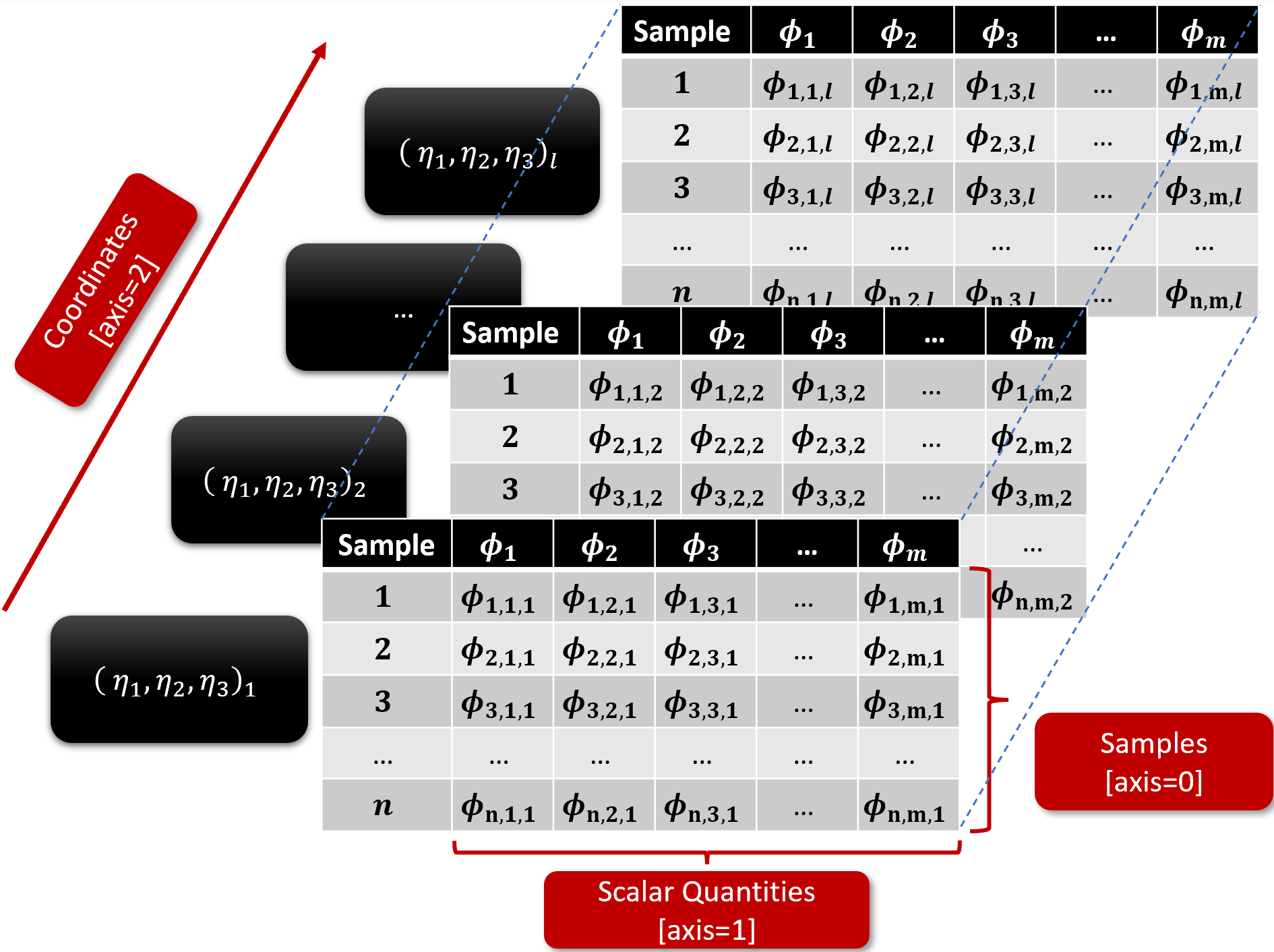}
    \caption{Data standard at import. 3D tensor, shape $(n,m,l)$. Prior to modeling pipeline ingestion, data should be in form \texttt{(samples,scalar\_values,coordinates)} or $(n,m,l)$. $\eta$ is a notional coordinate, often ($\eta_1,\eta_2,\eta_3$) is ($x,y,z$). $\phi$ here represents a scalar value at each coordinate, either an input parameter or a quantity of interest (QoI). A ``sample'' is a single instance of the measured experiment (for CFD, this would be one CFD snapshot).}
    \label{fig:data_standard}
\end{figure}

Two datasets were used to validate the modeling framework: a multi-fidelity dataset, presented in~\cite{korenyi2024josr}, and a panel methods dataset generated with CBAERO (Configuration-Based Aerodynamics), post-processed to include only the aerodynamic coefficients. This work establishes a standard data format, which is essential for effective and reproducible surrogate modeling. The data format is structured as follows: \texttt{(samples,scalar\_values,coordinates)}, or $(n,m,l)$. The data must be imported as a 3D tensor, as shown in Fig.~\ref{fig:data_standard}. Examples of what this looks like for one of the studied datasets can be found in Fig.~\ref{fig:US3D_data_example}. For each fidelity level, there will be one input data tensor and one output data tensor. The shapes of these two tensors need not match (and almost always will not match), but the number of samples must match ($n_{input} = n_{output}$). For the multi-fidelity dataset used in this study (discussed in Sec.~\ref{subsubsec:dataset2}), $n$ is the number of CFD cases, $m$ represents the number of fluid scalar variables (such as heat flux or pressure), and $l$ denotes the number of grid points. Generally, for a multi-fidelity surrogate modeling problem, there will be a many more low-fidelity training points ($n_{LF}$) than high-fidelity training points ($n_{HF}$): $n_{LF}>>n_{HF}$.

For certain engineering problems, exploring techniques that retain the spatial structure of the data is crucial. A limitation of the presented data format is that it requires consistent size distribution or mesh for each quantity of interest; data at each grid point must be uniform, and the model must be trained on equivalent meshes. If a model is trained on a single mesh across all flight conditions, generating model predictions on a different mesh/grid necessitates either retraining the model or performing interpolation between grids, which can introduce errors and require additional post-processing. The presented 3D tensor construct assumes a value at each coordinate point, using a uniform coordinate system for each sample. Modifications are needed to study datasets with varying coordinates (i.e., varying mesh resolution), using constructs like the deepXDE~\cite{lu2021deepxde} PointCloud object.

\begin{figure}[h!]
    \centering
    \includegraphics[width=0.7\textwidth]{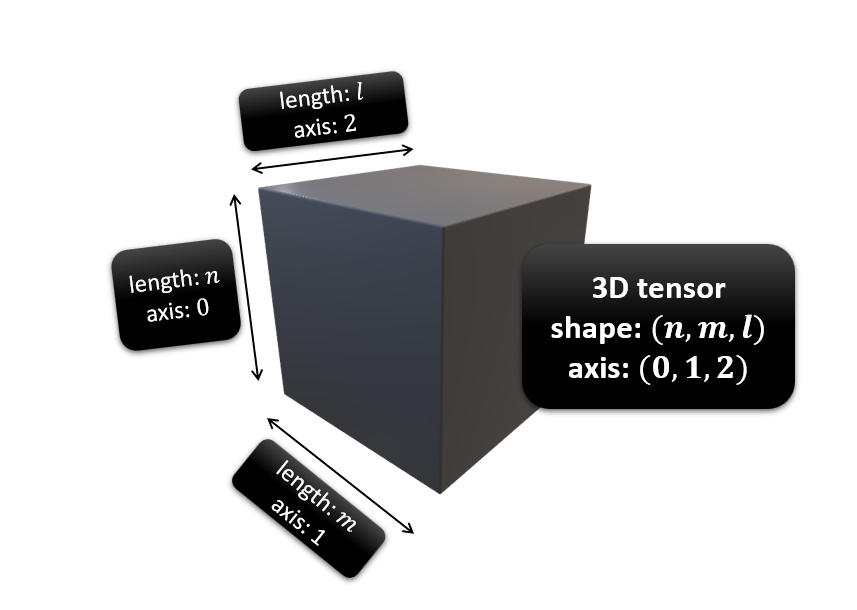}
    \caption{3D tensor visualization, shape (n,m,l).}
    \label{fig:3D_cube}
\end{figure}

  \begin{minipage}{\textwidth}
  \begin{minipage}[b]{0.49\textwidth}
    \centering
    \begin{tabular}{c|cccc}\hline
        \toprule        
        Sample & $T_W$ & $\rho_\infty$ & $T_\infty$ & $u_\infty$ \\
        \midrule
        1 & 450 & 0.08 & 129 & 1284 \\
        2 & 339 & 0.47 & 162 & 1871 \\
        ... & ... & ... & ... & ... \\
        400 & 350 & 0.37 & 200 & 1610 \\
        \bottomrule
      \end{tabular}
      \captionof{table}{Input data example. Shape: (400, 4, 1)}
  \end{minipage}
  \hfill
  \begin{minipage}[b]{0.49\textwidth}
    \centering
    \includegraphics[width=\textwidth]{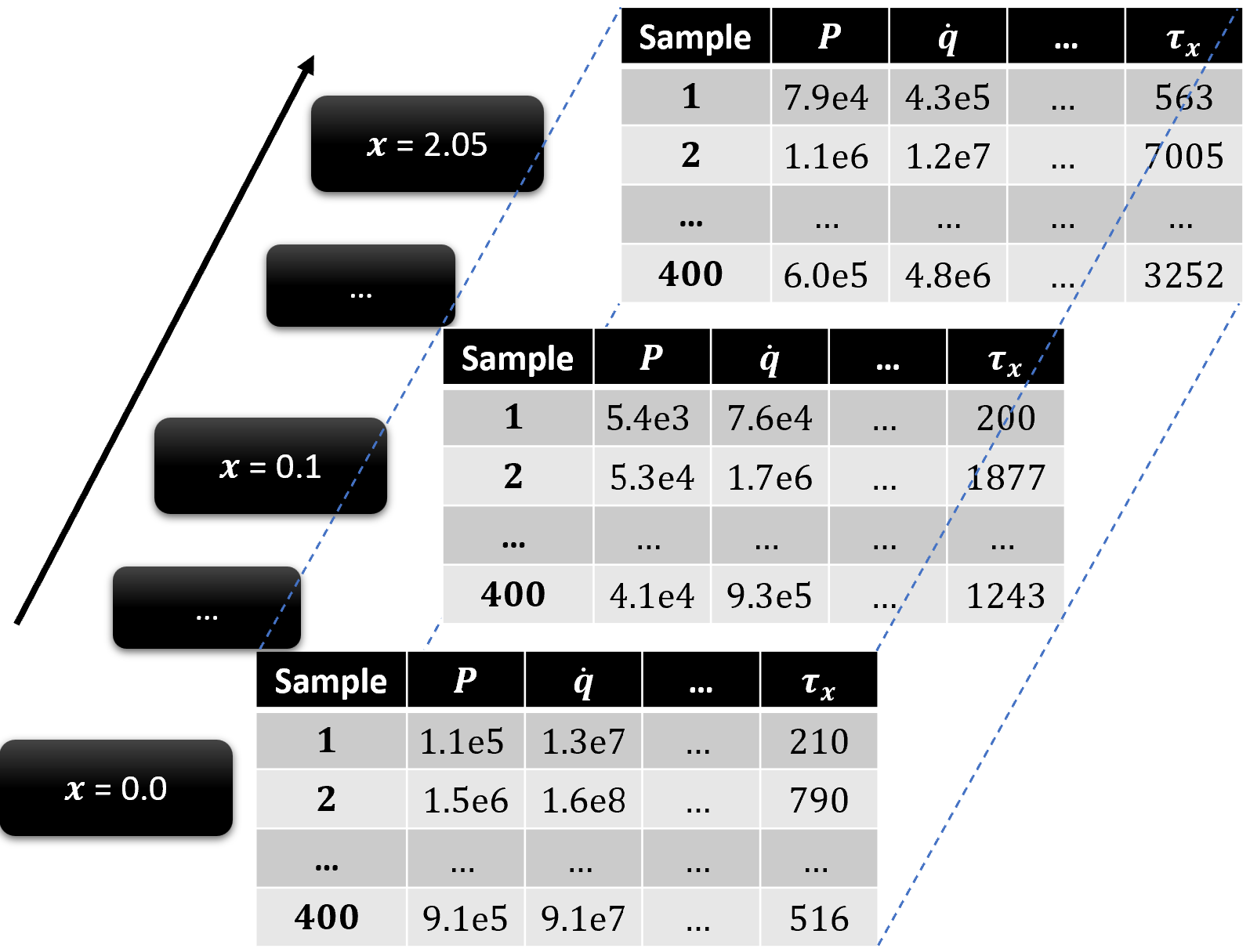}
    \captionof{figure}{Output data example. Shape: (400, 7, 1828)}\label{fig:US3D_data_example}
    \end{minipage}
  \end{minipage}

Despite these challenges, data-driven modeling with no explicit consideration for physical laws is effective for many engineering problems. Although some criticize the "naive" machine learning approach, it has proven successful in numerous aerospace applications~\cite{dreyer2017multi,Dreyer2021,korenyi2024josr,reasor_2021,dreyer2021dissertation,forrester2007multi,vanderwyst2016computationally, SMT2019,saves2024smt, queipo2005surrogate}. Specific problems, such as those involving shock fitting with varying meshes for each flight condition, incorporating scale-resolving or time-varying data, or integrating very limited/sparse experimental or flight test data, may require the more sophisticated methods discussed in Sec.~\ref{sec:future_work}.

\subsubsection{Dataset 1}\label{subsubsec:dataset1}
Dataset 1 is generated using CBAERO, a low-fidelity aerodynamic modeling tool. Due to its low computational expense, CBAERO allows for rapid solution evaluations without the need for high-performance computing resources. The dataset was generated solely to test the presented surrogate modeling framework on a previously unseen dataset, as Dataset 2 (found in Sec.~\ref{subsubsec:dataset2}) was used in the author's previous work. Input parameters are angle of attack ($\alpha$), angle of sideslip ($\beta$), altitude and Mach number. Quantities of interest (QoI) are aerodynamic coefficients, to include coefficient of lift ($C_L$), coefficient of drag ($C_D$), moment coefficient ($C_M$) and others. The dataset includes 4940 total steady-state simulations, each at different operating conditions. From the NASA website~\cite{kinney2004aero}: ``CBAERO is a software tool for the prediction of the conceptual aero-thermodynamic environments of aerospace configurations. The vehicle geometry is defined using unstructured, triangulated surface meshes. For supersonic and hypersonic Mach numbers, various independent panel methods are coupled with the streamline tracing formulation, attachment line detection methods, and stagnation-attachment line heating models to define the viscous aero-thermal environment." 

Pseudo-code for an example data-cleaning script can be found in Algorithm \ref{alg:process_hgv_data}, which converts CBAERO aerodynamic coefficients output data into a formatted aerodynamic table compatible with this modeling framework. When using a new data source, a new post-processing script must be created to convert the data into the prescribed format. For many problems, this preprocessing script is the only custom work required to use this framework for surrogate modeling. Once the data is formatted correctly, the framework can automatically handle all preprocessing, scaling, splitting, training, and testing tasks.

\begin{figure}[H]
    \centering
    \includegraphics[width=0.6\textwidth]{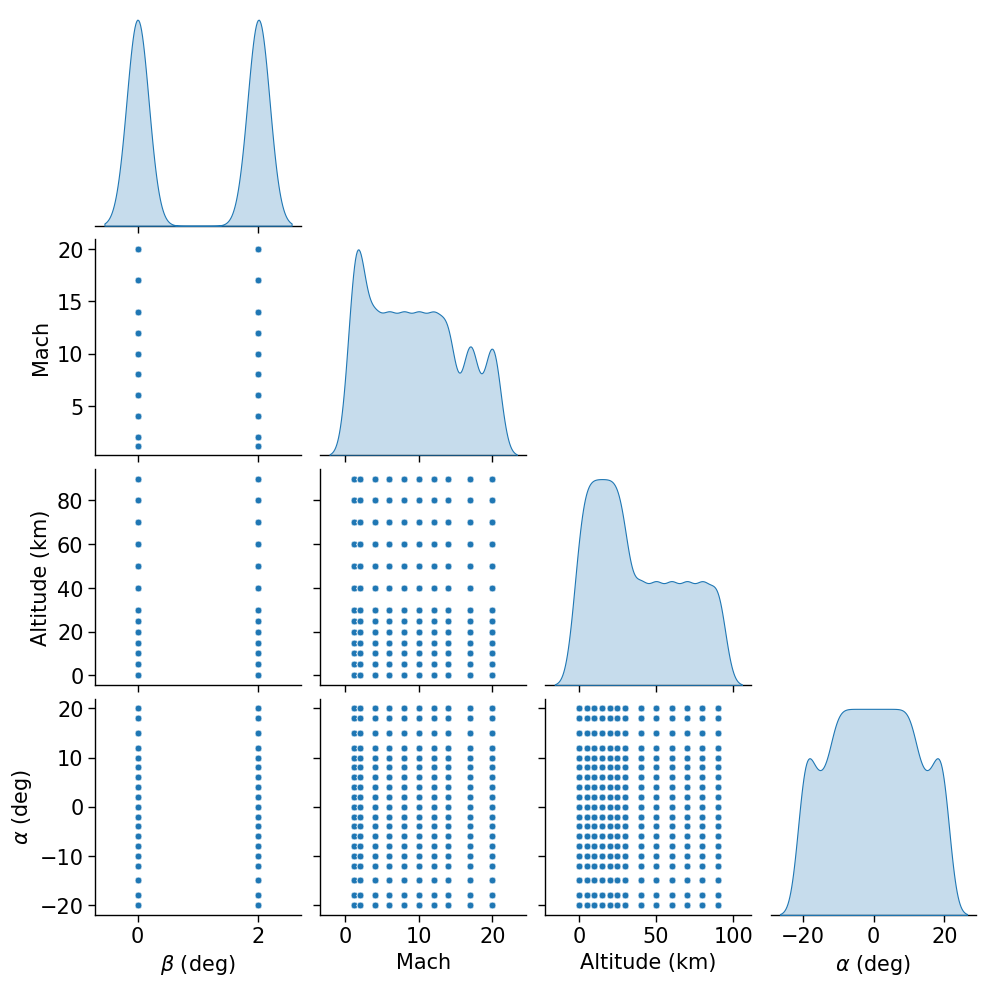}
    \caption{Input space pairplot for CBAERO dataset. Pairplots show pairwise relationships between parameters, in this case showing the input parameter space coverage.}
    \label{fig:cbaero_input_space}
\end{figure}

  \begin{minipage}{\textwidth}
  \begin{minipage}[b]{0.49\textwidth}
    \centering
    \includegraphics[width=\textwidth]{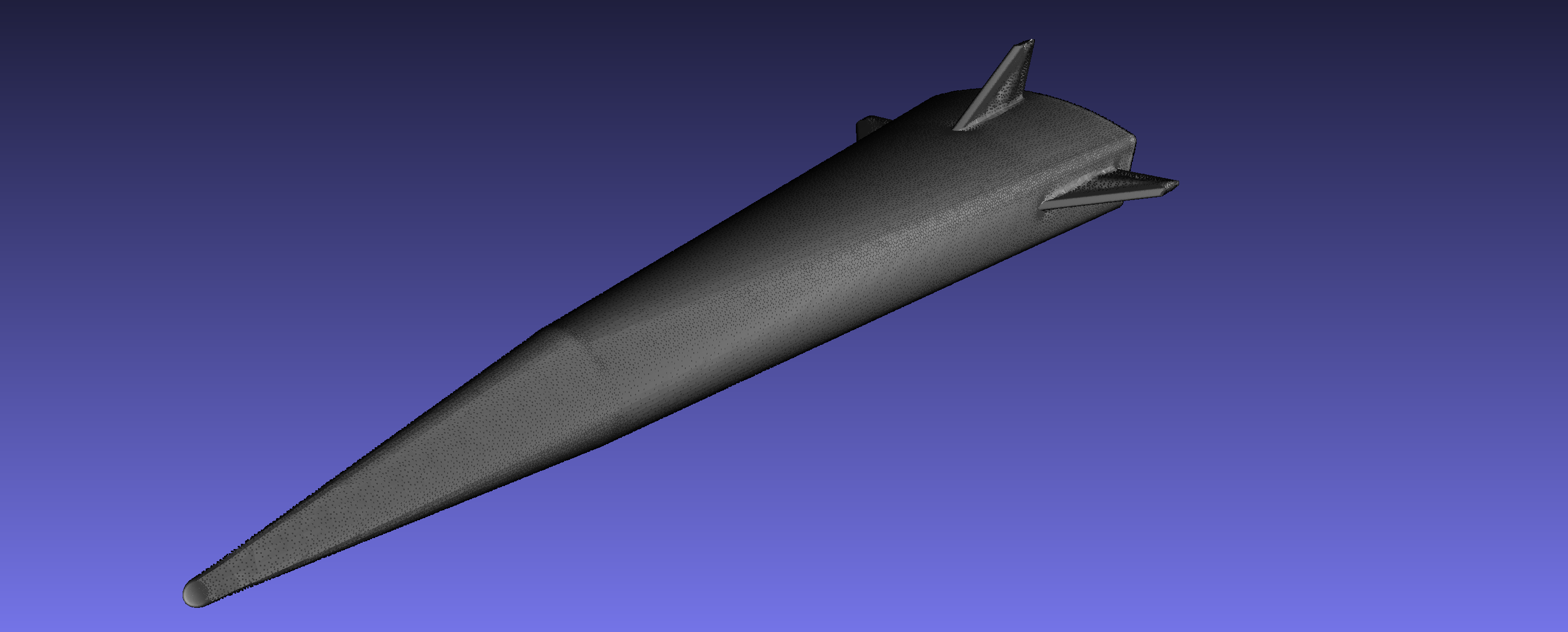}
    \captionof{figure}{Vehicle geometry}
    \vspace*{\fill}
  \end{minipage}
  \hfill
  \begin{minipage}[b]{0.49\textwidth}
    \centering
    \begin{tabular}{ccc}\hline
        \toprule
        \textbf{Lower Limit} & \textbf{Variable} & \textbf{Upper Limit} \\
        \midrule
        -20 & $\alpha$ \, [deg] & 20 \\
        0 & $\beta$ \, [deg] & 2 \\
        0 & Altitude \, [km] & 90 \\
        1.2 & Mach & 20 \\
        \bottomrule
      \end{tabular}
      \captionof{table}{Dataset 1 operating conditions}
      \vspace*{\fill}
    \end{minipage}
  \end{minipage}

\subsubsection{Dataset 2}\label{subsubsec:dataset2}
Dataset 2 is generated using inviscid oblique shock relations (low-fidelity) and RANS CFD solutions (high-fidelity). Detailed information regarding this multi-fidelity dataset can be found in~\cite{korenyi2023thesis,korenyi2024josr}. Input parameters are the isothermal wall temperature ($T_w$), freestream density ($\rho_\infty$), freestream temperature ($T_\infty$), and freestream velocity ($u_\infty$). QoI for this dataset are the wall heat flux ($\dot{q}_w$), wall pressure ($P_w$), and wall shear stress ($\tau_w$). These parameters are essential for understanding the aerodynamic and thermal performance of high-speed vehicles. The pressure and shear stress on the vehicle's surface is used to calculate aerodynamic coefficients of (lift coefficient, $C_L$ and drag coefficient, $C_D$), which are used to assess the vehicle's aerodynamic performance. The heat flux on the vehicle's surface is used for conducting thermal analysis, as aerodynamic heating presents a design challenge in the high-speed flight regime. Over-designing for the high heating environment can result in a heavier-than-necessary flight vehicle, thereby limiting range and performance of the flight vehicle or program cancellation due to inability to close design loop. As the vehicle is axisymmetric, only a quasi-1D distribution along the surface of the cone was considered. 

\begin{figure}[H]
	 \begin{subfigmatrix}{2}
	 	  \subfigure[Flight vehicle shape geometry (test article from an experimental study by \cite{wadhams2014comparison} (figure credit Wadhams et al. 2014).]{\includegraphics[width=0.49\textwidth]{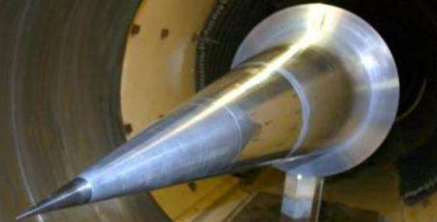}}
	 	  \subfigure[Computational fluid domain for RANS solutions (Credit: Mr. Jon Willems, Jr., figure from \cite{korenyi2024josr})\label{subfig:computational_RANS_domain}]{\includegraphics[width=0.49\textwidth]{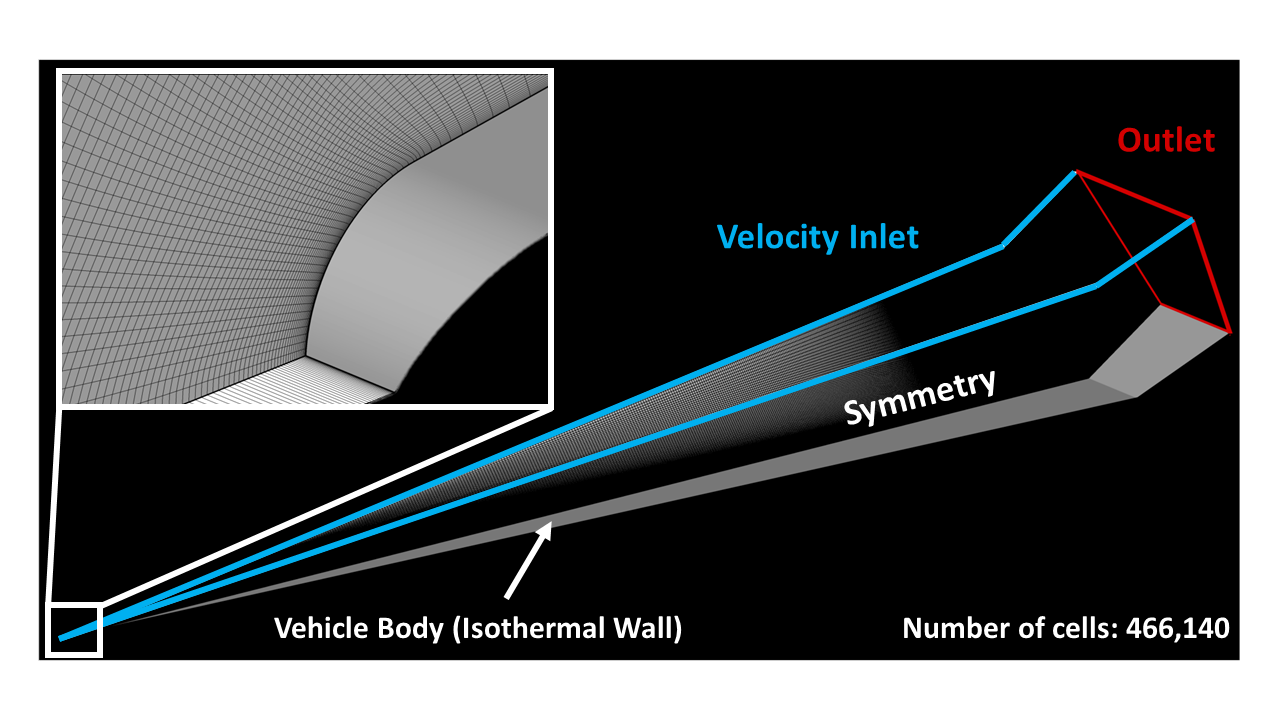}}
	 \end{subfigmatrix}
	 \caption{Geometry and domain}
	 \label{fig:geometry_and_domain}
\end{figure}

\begin{figure}[H]
    \centering
    \includegraphics[width=0.9\textwidth]{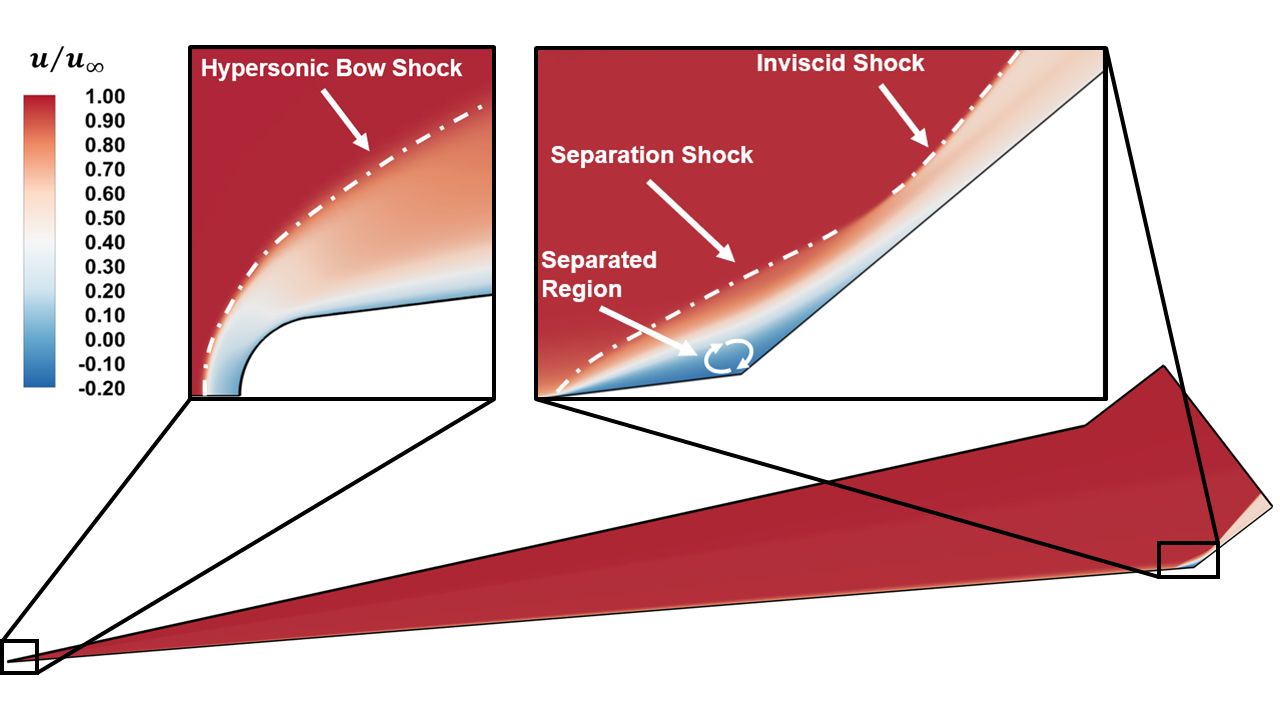}
    \caption{Representative US3D solution on cone flare (Credit: Mr. Jon Willems, Jr., figure from \cite{korenyi2024josr})}
    \label{fig:us3d_sol}
\end{figure}

\begin{figure}[H]
    \centering
    \includegraphics[width=0.6\textwidth]{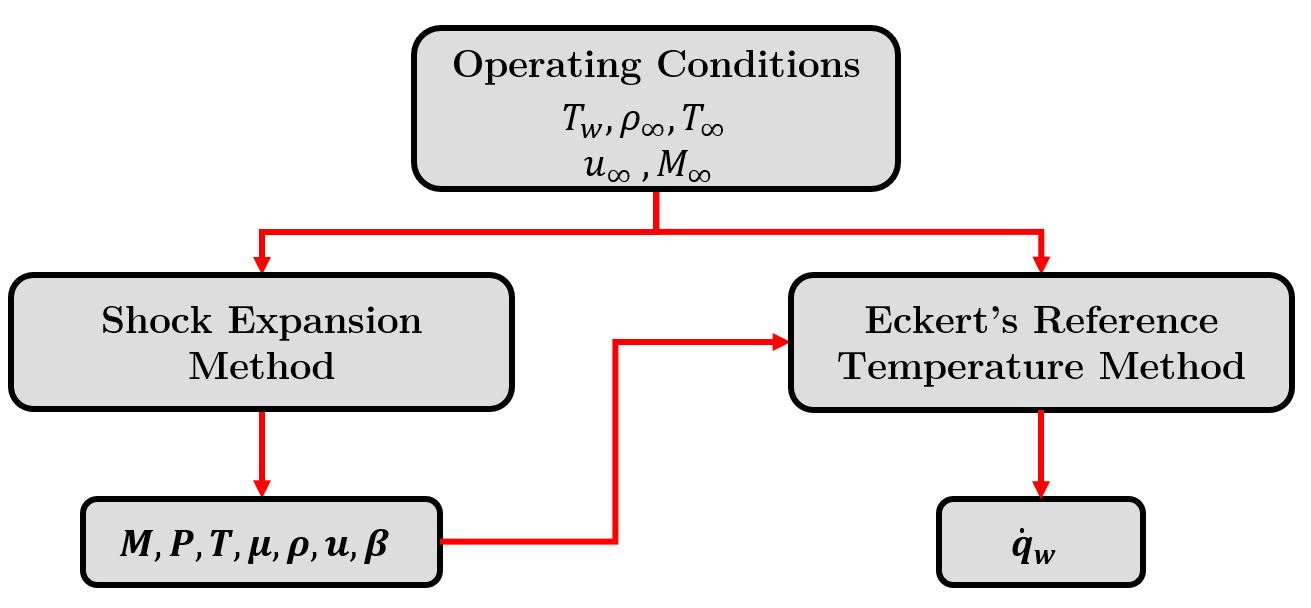}
    \caption{Low-fidelity data generation workflow. Shock-expansion (S-E) pressure values are calculated based on flight conditions, then surface heating is calculated using S-E pressure, isothermal wall temperature, and flight conditions. }
    \label{fig:lowFidelityDataGenerationWorkflow}
\end{figure}

The high fidelity training data was generated using the Reynolds-Averaged Navier Stokes (RANS) computational fluid dynamics (CFD) solver, US3D\cite{candlerUS3D2015}, a popular high-speed flight code used broadly by government, industry, and academia. The one-equation Spallart--Allmaras turbulence model is used. The axisymmetric computational domain for the US3D RANS simulations is shown in Fig.~\ref{subfig:computational_RANS_domain}. Of note, while US3D is capable of accounting for real-gas effects, these effects were not considered in this study.  The high-fidelity data was computed at 400 sampling points in the parameter space.  A representative result is shown in Fig.~\ref{fig:us3d_sol}. The low-fidelity data was generated at 800 sampling points in the parameter space.  Two methods were used to generate the low fidelity data: shock-expansion method~\cite{anderson_2019} to predict surface pressure, and Eckert's reference temperature method~\cite{Eckert_1956} to predict surface heat transfer. The low-fidelity data generation process is shown in Fig.~\ref{fig:lowFidelityDataGenerationWorkflow}. This dataset was generated by Dr. Daniel Reasor and Mr. Jon Willems, Jr.

\begin{figure}[H]
	 \begin{subfigmatrix}{2}
	 	  \subfigure[Low fidelity data input space]{\includegraphics[width=0.49\textwidth]{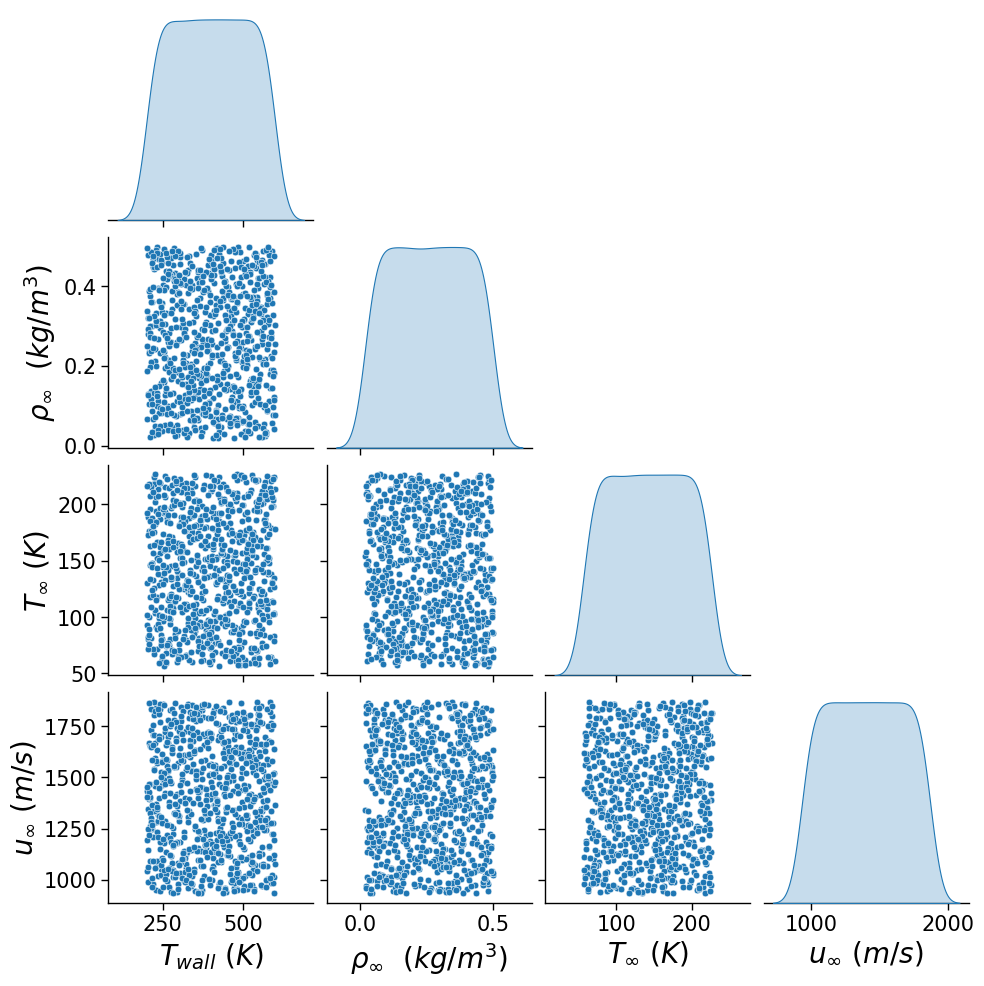}}
	 	  \subfigure[High fidelity data input space]{\includegraphics[width=0.49\textwidth]{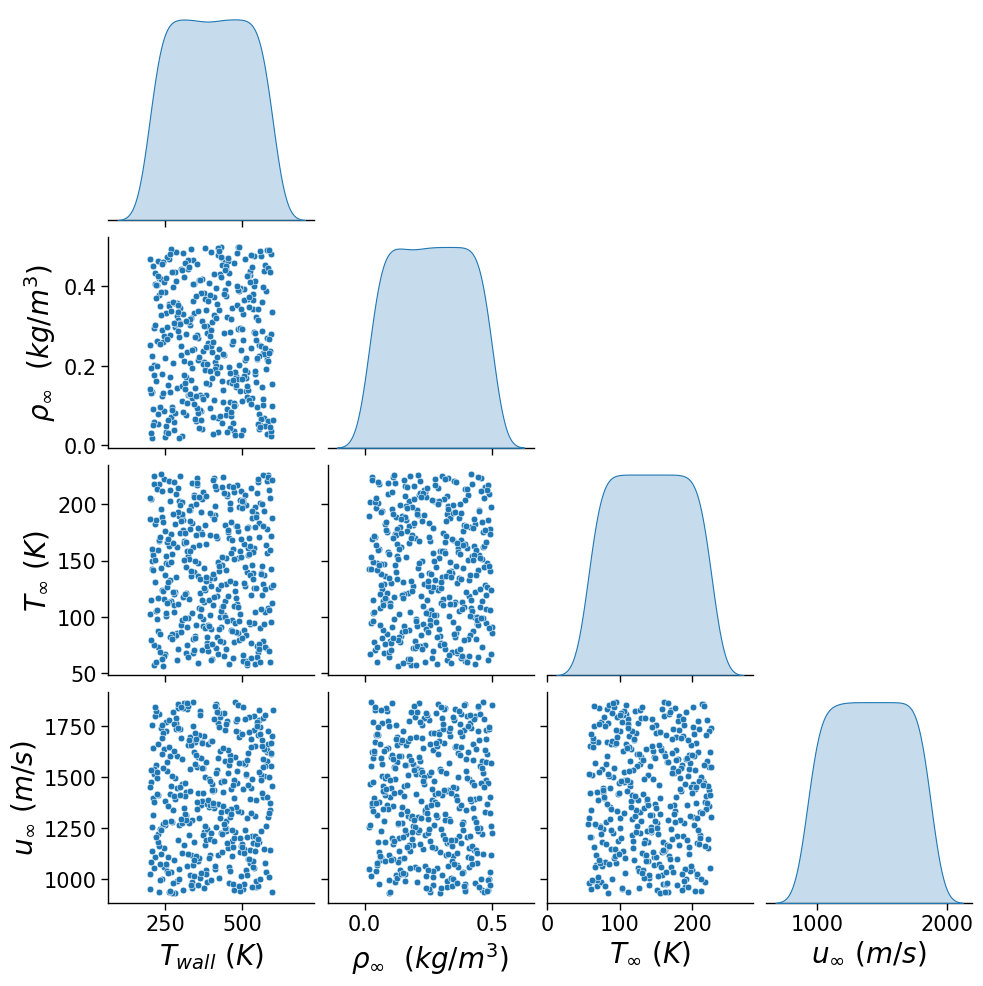}}
	 \end{subfigmatrix}
	 \caption{Input space pairplot for multi-fidelity dataset. Pairplots show pairwise relationships between parameters, in this case showing the input parameter space coverage.}
	 \label{fig:MF_pairplot}
\end{figure}

\subsection{Model Implementation}

The surrogate modeling framework used is inspired by Guo et al. 2022~\cite{GUO2022114378} and implicitly by Meng et al. and Motamed et al.~\cite{meng2020composite,motamed2020multi}. The framework is written in Python 3.9~\cite{python}. Pandas~\cite{Pandas_mckinney2010data} and NumPy~\cite{harris2020array} are used for storing and handling CFD data, while scikit-learn~\cite{pedregosa2011scikit} is used for data scaling, normalization, cross-validation, and implementing Gaussian process regression (GPR) models. TensorFlow~\cite{abadi2016tensorflow} with Keras API~\cite{chollet2015keras} is used to implement all neural network training and prediction tasks. The framework also incorporates uncertainty quantification (via scikit-learn GPR output variance) and hyperparameter tuning (written in Python).

Uncertainty quantification (UQ) is used for informing active sampling and assessing the reliability of model predictions. Model uncertainty is linked to model accuracy~\cite{korenyi2024josr}, and thus it provides valuable insights into model trustworthiness when generating predictions on new data. There is a wide body of literature on machine learning model UQ~\cite{pmlr-v48-gal16,abdar2021review}. UQ analysis is integrated into the presented framework and will be used in future work to inform active sampling strategies. The UQ incorporated in this framework is specifically for GPR models, using the standard deviation and mean values generated by these models.

Hyperparameter tuning is required for building effective surrogate models but can be challenging for non-experts. This framework employs a subset of hyperparameters that have proven effective for high-speed flight aerodynamic data, and performs hyperparameter sweeps over this subset. Although not the most mathematically rigorous process, it is straightforward, time efficient, and resource efficient. Engineers can easily modify or expand the default hyperparameter search space if desired. Tuning the scikit-learn implementation of GPR is straightforward, with the primary decision being the choice of kernel. The model will automatically tune the kernel hyperparameters during model fit by finding the values that maximize the log-marginal likelihood. The radial basis function (RBF) kernel is often the ``standard" GPR kernel, while the Matern kernel offers more flexibility (through varying the kernel's $\nu$ parameter, which is user-defined and not fit during training). More on kernels can be found in the scikit-learn~\cite{pedregosa2011scikit} \href{https://scikit-learn.org/stable/modules/generated/sklearn.gaussian_process.kernels.Matern.html}{documentation} or in the \href{https://www.cs.toronto.edu/~duvenaud/cookbook/}{Kernel Cookbook}~\cite{duvenaud_2014}. Tuning shallow DNNs, which are the primary NNs considered in this framework, depends largely on the number of layers and the width of the layers. Most tuning efforts focus on finding the Pareto optimal neural network size, utilizing a vectorized \texttt{for} loop to test different layer numbers and widths. For other hyperparameters more suited for optimized hyperparameter search, KerasTuner~\cite{omalley2019kerastuner} is used. 

\section{Results}
\begin{figure}[H]
	 \begin{subfigmatrix}{2}
	 	  \subfigure[Both the neural network-based and GPR-based models accurately predicted the aerodynamic coefficients in the CBAERO test dataset (data withheld during training) with $R^2$ greater than 0.99. \label{subfig:CBAERO_QOI_normed}]{\includegraphics[width=0.49\textwidth]{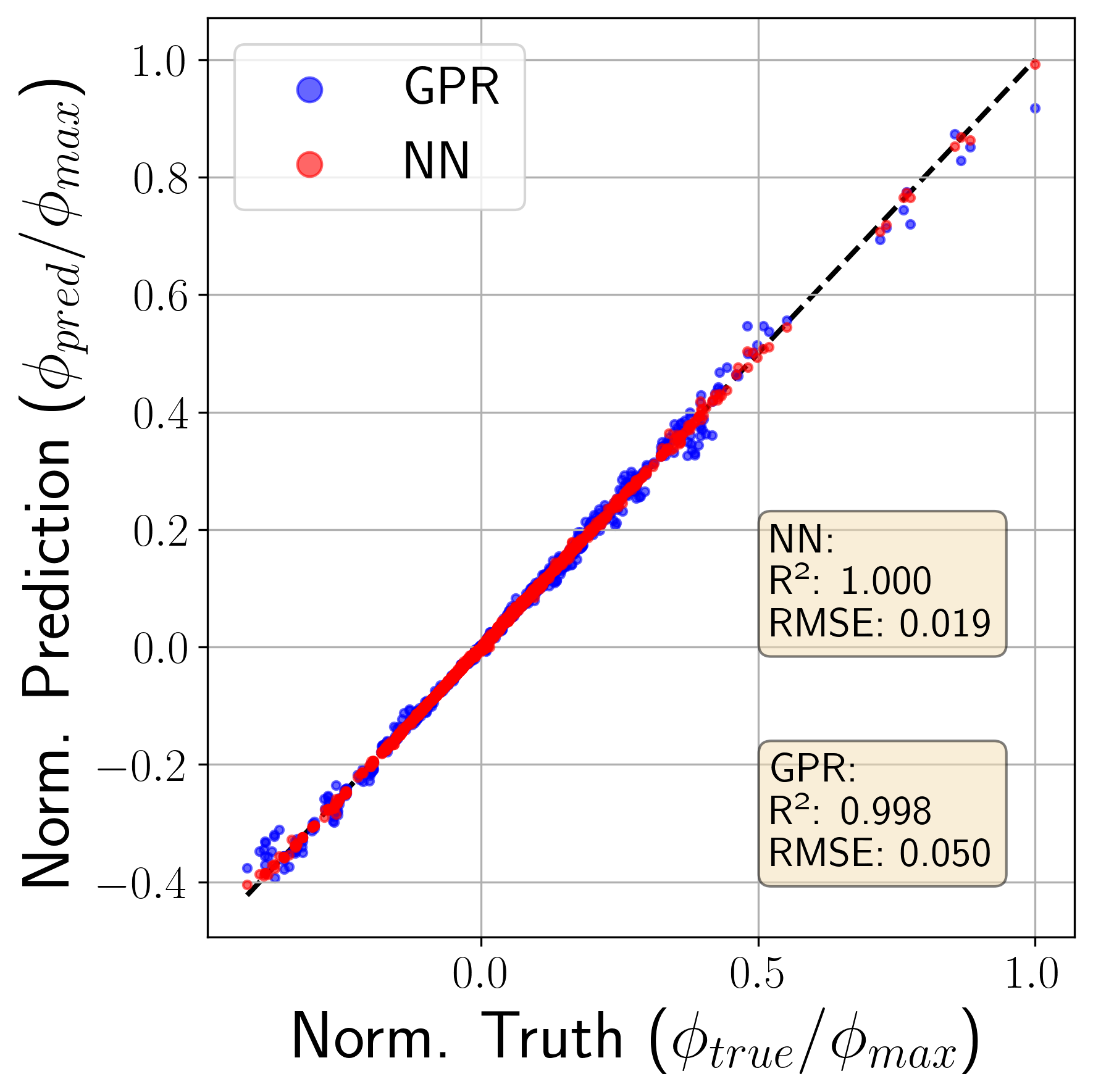}}
	 	  \subfigure[Both the neural network-based and GPR-based models predicted surface quantities in the US3D test dataset (data withheld during training) with $R^2$ greater than 0.99. ]{\includegraphics[width=0.49\textwidth]{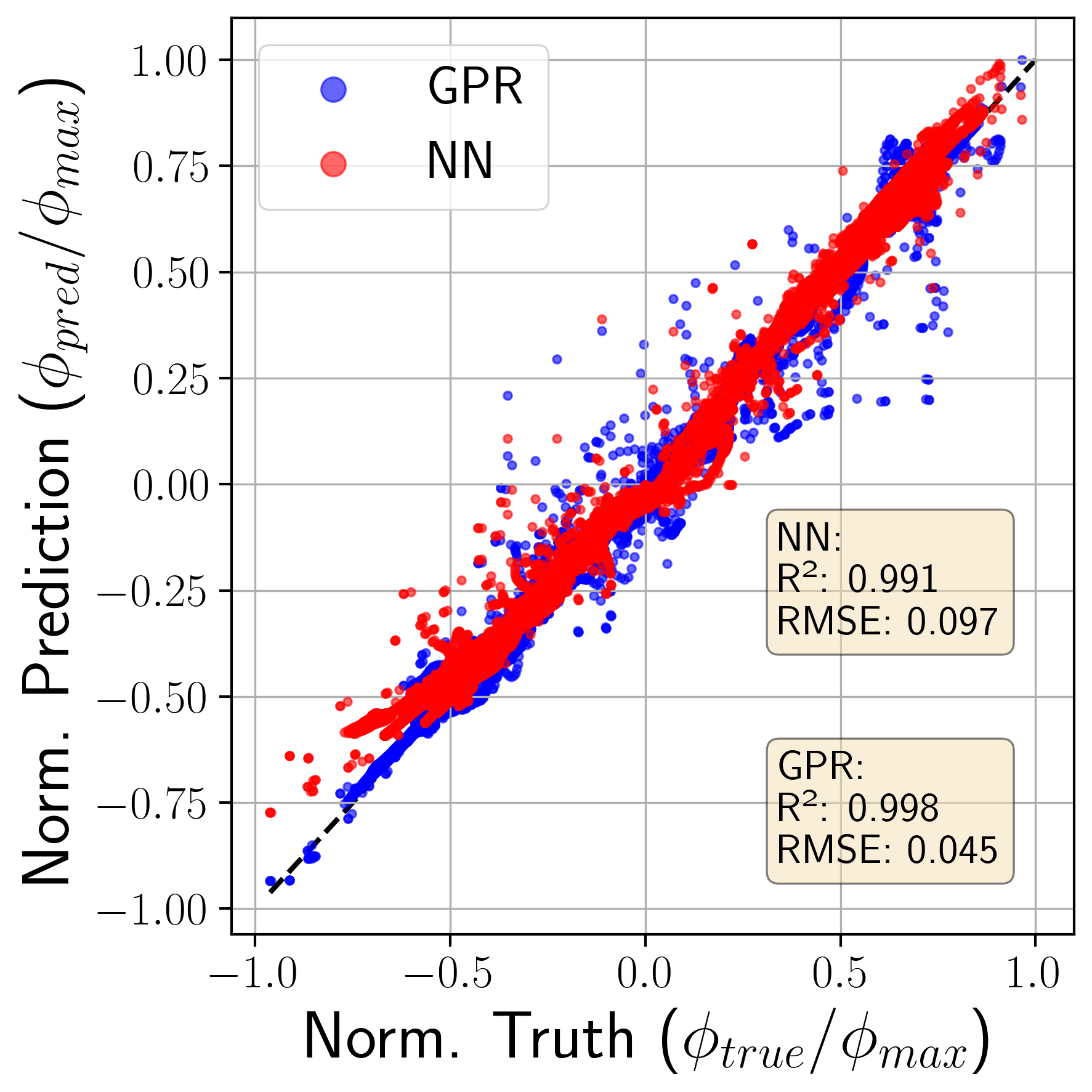}}
	 \end{subfigmatrix}
	 \caption{One to one scatter plots and model accuracy summary statistics}
	 \label{fig:CBAERO_model_accuracy}
\end{figure}

The primary deliverable of this project is a usable, open-source, modular, and straightforward end-to-end pipeline for multi-fidelity surrogate modeling.  Performance results are provided for two representative problems: a single-fidelity problem using CBAERO for high-speed flight vehicles (detailed in Sec.\ref{subsubsec:dataset1}) and a multi-fidelity problem combining panel methods with RANS (detailed in Sec.\ref{subsubsec:dataset2}). The multi-fidelity example replicates results from~\cite{korenyi2024josr, korenyi2023thesis}, demonstrating that the framework can reproduce the findings of these studies and serves as an initial validation of the framework's functionality. The framework also demonstrates good performance on a new dataset. This is illustrated in Fig.~\ref{fig:CBAERO_model_accuracy}, showing results for all normalized QoI for both datasets. Normalizing the predictions allows for all QoIs to be displayed on the same figure. The primary goal of this study is not to surpass performance benchmarks but to demonstrate that the code can effectively train and evaluate surrogate models across different datasets. The performance results are intended to validate that the framework is functional and reliable. The key contribution of this paper is the presentation of a usable, simple framework that engineers can seamlessly implement today. Additional validation and testing will continue over the next year at the author's home unit.

\begin{figure}[H]
	 \begin{subfigmatrix}{2}
                \subfigure[Surface heat flux distributions from the test set compared to model predictions]{\includegraphics[width=0.49\textwidth]{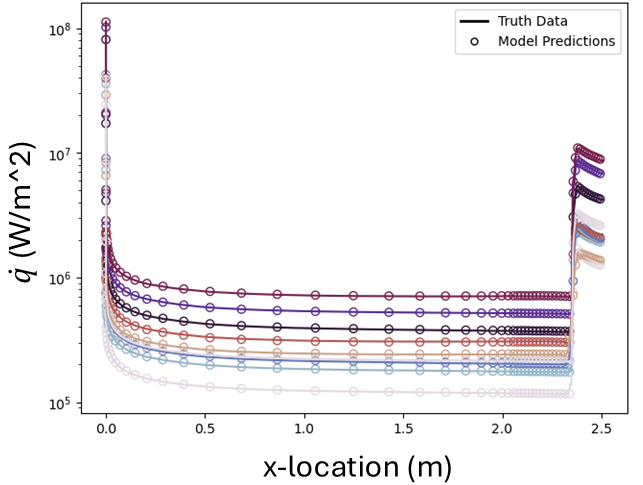}}
	 	  \subfigure[Surface pressure distributions from the test set compared to model predictions]{\includegraphics[width=0.49\textwidth]{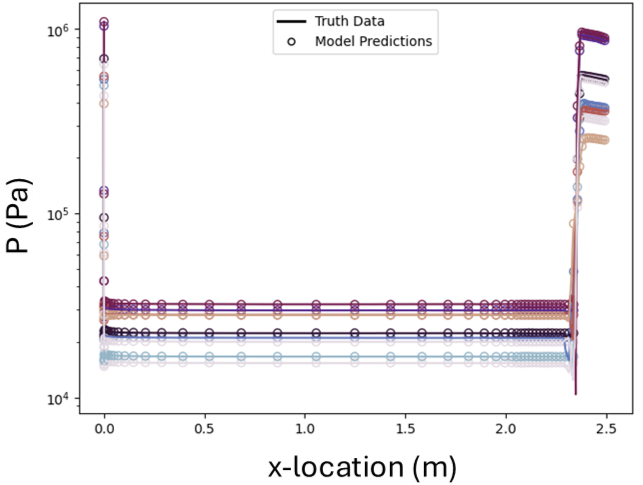}}
	 \end{subfigmatrix}
	 \caption{Multi-fidelity GPR-based model single case predictions for randomly selected samples from test dataset. Indicates strong model performance, matches results from prior work}
	 \label{fig:US3D_results_random_test_cases}
\end{figure}

\begin{figure}[H]
	 \begin{subfigmatrix}{2}
                \subfigure[$\alpha$ vs. $C_L/C_D$: Actual vs. predicted values across various angles of attack. Color gradient indicates altitude. Model predictions closely align with actual values.]{\includegraphics[width=0.49\textwidth]{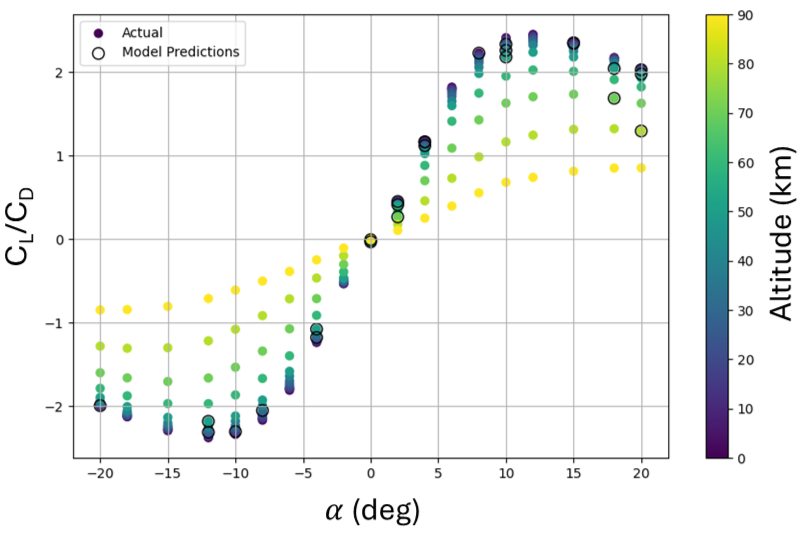}}
	 	  \subfigure[Altitude vs. $C_L/C_D$: Actual vs. predicted values across altitude values. Color gradient indicates angle of attack ($\alpha$). Model predictions closely align with actual values.]{\includegraphics[width=0.49\textwidth]{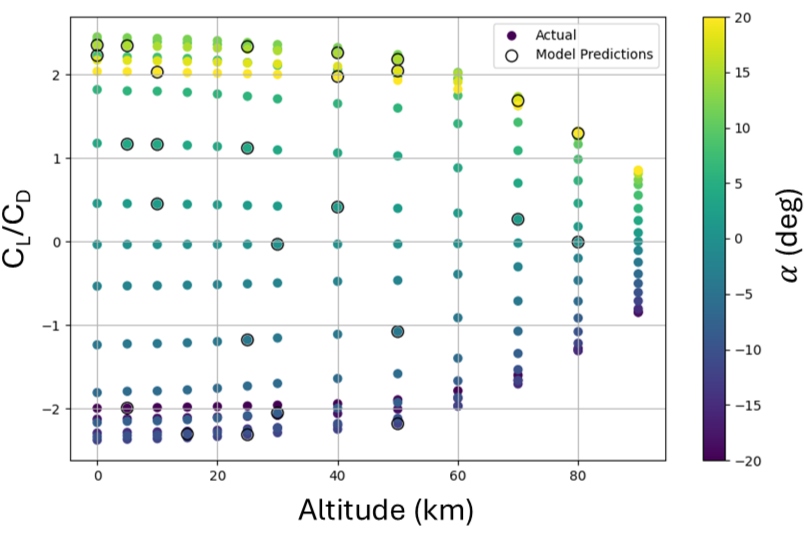}}
	 \end{subfigmatrix}
	 \caption{2D slices of the results shown in Fig.~\ref{fig:3D_model_performance}.}
	 \label{fig:2D_slice_model_performance}
\end{figure}

\begin{figure}[H]
    \centering
    \includegraphics[width=0.5\textwidth]{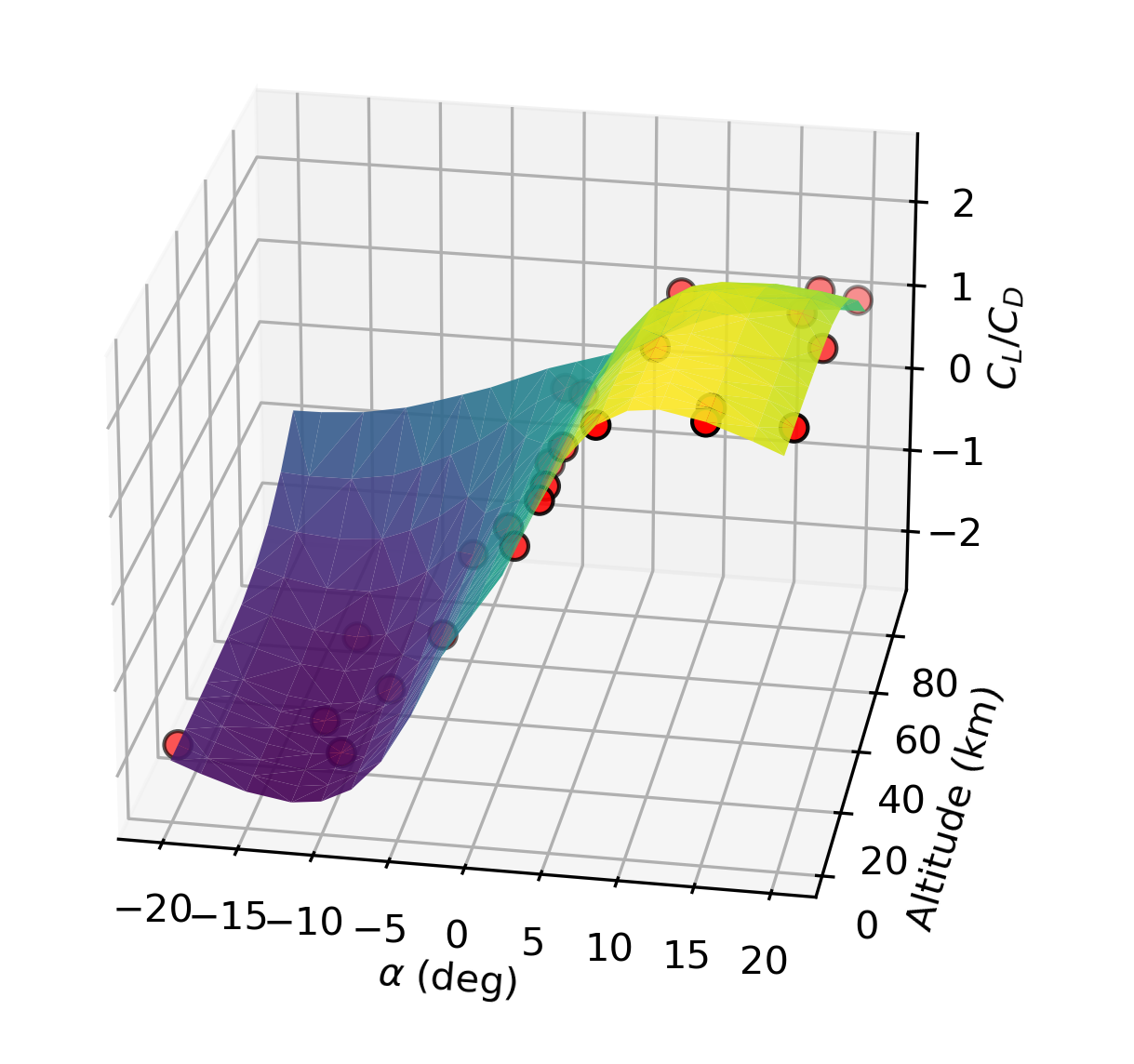}
    \caption{3D response surface, showing $C_L/C_D$ dependence on angle of attack ($\alpha$) and altitude. Model predictions indicated by the red points, which all lie on the surface. Visual representation of high model accuracy. More granularity in Fig.~\ref{fig:2D_slice_model_performance}}
    \label{fig:3D_model_performance}
\end{figure}

Representative data from both datasets are presented in Figures~\ref{fig:US3D_results_random_test_cases}, ~\ref{fig:2D_slice_model_performance}, and \ref{fig:3D_model_performance} to provide further insight into the model predictions. Notably, the model inference times exceed 1000 Hz, making these models suitable for online optimization and evaluation. Further details on model evaluation times for the US3D dataset are available in~\cite{korenyi2024josr}.

Additionally, high-speed flight vehicle aerodynamic coefficients for a large flight envelope were accurately modeled without full flow field, sub-domain, or surface values. Often, surrogate models for high-speed vehicles train on and predict sub-domain or surface values, requiring post-processing to obtain aerodynamic coefficients. By training directly on aerodynamic coefficients, computational costs were reduced, resulting in compact models with fewer parameters. This efficiency is attributed to the dense sampling of the large training dataset, minimizing the interpolation required.

\section{Future Work}\label{sec:future_work}

This research will evolve in alignment with the field's progression towards context-aware machine learning. However, increasing complexity and intrusiveness can reduce flexibility. Many approaches involve encoding physics, but manually encoding governing equations can be challenging for practical engineering problems.


This work will be expanded through two research efforts, through the DAF-MIT AI Accelerator and the DAF-Stanford AI Studio. The author will continue to develop practical, user-facing functionality while establishing a larger-scale collaborative effort with top research faculty. This approach aims to enhance research efficiency, deconflict research focus, and ultimately deliver cutting-edge capabilities to the Department of Defense. 

The immediate steps to be executed in the near term:
\begin{itemize}
    \item \textbf{Incorporate optimized GPR:} Currently, the presented framework's only option for GPR is exact GPR, which becomes intractable for large datasets. Packages under consideration for optimized GPR include SMT GP~\cite{saves2024smt}, OILMM~\cite{bruinsma_scalable_2020}, and PyTorch GP~\cite{gardner2018gpytorch}.
    \item \textbf{Dimensionality reduction}: The first iteration of dimensionality reduction in this framework will be Proper Orthogonal Decomposition-Kriging (POD-Kriging). There are many examples of the effectiveness of POD-based dimensionality reduction~\cite{falkiewicz2011reduced, bostanabad_globally_2019, bouhlel2019gradient, SMT2019, barnett2023neural}.
    \item \textbf{Alternative modeling methodologies:} Exploring kNN regression for its simplicity and ease of tuning, and Bayesian Neural Networks to incorporate uncertainty quantification in DNNs, though the latter may come with increased training costs. 
    \item \textbf{Enhanced design of experiments (DoE):}   Incorporating trajectory-informed sampling~\cite{needels2024trajectory} to increase efficiency and decrease computational expense. Explore various active learning~\cite{kamath2022intelligent} methods, first using uncertainty quantification data for informing future sampling.
    \item \textbf{User interface (UI) / User experience (UX) improvement:} Develop a more streamlined user-facing script. Then, develop a limited-use GUI. Finally, integrate into a unified aerodynamic analysis framework. 
    \item \textbf{Optimization and dependency reduction:} Streamlining the framework to reduce its dependency on numerous, potentially heavy libraries. This process would benefit from a software engineer's expertise to optimize performance.
    \item \textbf{Evaluate Dockerized deployment:} Assessing the trade-offs of containerizing the application to simplify deployment to air-gapped systems.
    \item \textbf{Initial exploration of advanced modeling techniques:} Begin initial implementation and testing of advanced techniques such as Fourier neural operators (FNO), Physics-informed neural operators (PINOs), graph neural networks (GNNs), and physics-informed neural networks (PINNs). This exploratory work will lay the foundation for the more extensive research and development efforts planned through the DAF AI Accelerator and DAF AI Studio.

\end{itemize}

\section{Conclusion}
High-speed flight vehicle modeling is challenging and expensive. Currently, the government's production-ready capabilities lag behind the cutting edge research technologies available in academia. This project delivered a minimum viable product (MVP) to the Air Force incorporating technologies as recent as 2022, demonstrating the feasibility and potential of advanced surrogate modeling techniques. The utility of the framework was demonstrated with two datasets, both representative of practical high-speed flight modeling problems. This framework is also broadly applicable to high-dimensional, non-linear regression type scientific machine learning problems, provided the user has a dataset with sufficient correlation between the input and output data. However, many challenges remain. Recent advancements in physics-informed machine learning, such as physics-informed neural operators, PINNs, and physics-enhanced deep surrogates, have shown promise in solving outstanding research problems like modeling varying geometries, handling varying meshes, and generating predictions at or beyond the edges of the input parameter space (i.e., extrapolation). Serious and immediate investment in continued research is crucial. Collaborations with Stanford University and MIT, facilitated through the DAF AI Accelerator and DAF AI Studio, should be continued. These partnerships will help ensure the US government remains at the forefront of high-speed flight vehicle modeling and simulation.

\section*{Acknowledgments}
The author is grateful to his home unit and the DAF-MIT AI Accelerator for this tremendous opportunity. Thank you to the Cohort 10 Phantoms for being world class. Special thanks to Rachel Price, Greg Search, and Peyton Cooper for the diligent technical editing and hours of discussion related to this work.  Dr. Daniel Reasor and Mr. Jon Willems, Jr. are responsible for generating the RANS data in Dataset 1, and the author's colleagues at his home unit are responsible for generating the data in Dataset 2. Special thanks to Professor Jack McNamara for his mentorship and guidance during my time at Ohio State and beyond--his advisement was invaluable. 

\appendix

\section*{Appendix}

\section{Modeling Techniques}\label{app:modeling_techniques}

This section provides an abbreviated summary of the detailed information contained in the author's thesis~\cite{korenyi2023thesis} and a recent AIAA paper~\cite{korenyi2024josr}. For a more comprehensive and rigorous treatment of these topics, please refer to those sources.

\subsection{Gaussian Process Regression}
Gaussian Process Regression, is a statistical interpolation process originally suggested by a Danie Krige\cite{krige1951statistical} and refined by Georges Matheron\cite{matheron1963principles}.  The underlying mathematics of Gaussian process was used at least as early as Wiener~\cite{wiener1949extrapolation} and Kolmogorov~\cite{kolmogoroff1941interpolation} in the 1940s. A Gaussian process is represented by it's mean, $m(\bm{x})$, and covariance, $k(\bm{x},\bm{x'})$:

\begin{equation}
    f(\bm{x}) \sim \mathcal{GP}(m(\bm{x}),\,k(\bm{x},\bm{x'}))
\end{equation}

The presented framework uses scikit-learn's GPR module, which follows Algorithm 2.1 from Rasmussen and Williams~\cite{Rasmussen2006}. More on Gaussian process for machine learning can be found in Rasmussen and William's foundational text~\cite{Rasmussen2006}.
\begin{algorithm}[H]
\caption{Gaussian Process Regression}\label{alg:GPR_rasmussen}
\begin{algorithmic}[1]
\Procedure{GPR}{$X$, $y$, $X_*$}
    \State \textbf{Input:} Training input $X$, target $y$, covariance function (or kernel) $K$, noise level $\sigma_n^2$, test input $x_*$
    \State \textbf{Output:} Mean predictions $\Bar{f_*}$, variance $\mathbb{V}[f_*]$, log marginal likelihood $\log p(y|X,\theta)$, 
    
    \State Compute covariance matrix $K = K(X, X)$
    \State Compute Cholesky decomposition $L = \texttt{cholesky}(K + \sigma_n^2 I)$
    
    \State Solve for $\bm{\alpha}$: $\bm{\alpha} = L^\top \backslash (L \backslash y)$
    
    \State Compute covariance between test and training points $K_* = K(X_*, X)$
    \State Compute mean predictions $\Bar{f_*} = K_*^\top \bm{\alpha}$
    
    \State Solve for $\mathbf{v}$: $\mathbf{v} = L \backslash K_*$

    \State Compute covariance between test points $K_{**} = K(X_*, X_*)$
    \State Compute variance $\mathbb{V}[f_*] = K_{**} - \mathbf{v}^\top \mathbf{v}$
    
    \State Compute log marginal likelihood: $\log p(y|X,\theta) = -\frac{1}{2} y^\top \bm{\alpha} - \sum_i \log L_{ii} - \frac{n}{2} \log 2\pi$

    \State \Return Mean prediction $\Bar{f_*}$, variance $\mathbb{V}[f_*]$, log marginal likelihood $\log p(y|X,\theta)$
\EndProcedure
\end{algorithmic}
\end{algorithm}

\subsection{Deep Neural Networks}
Artificial neural networks were first postulated by McCulloch and Pitts in 1943~\cite{mcculloch_pitts_1943} and first implemented by Rosenblatt in 1956~\cite{rosenblatt_1957}.  The artificial neuron is the building block used to construct an artificial neural network.  It works by receiving a signal, $x$, which is transformed using a linear combination of weights, $z(x,w) = w_0 + \sum_{i=1}^m w_i x_i$, where $w_i$ is a weights and $x_i$ is an input from a previous neuron.  Next, $z$ is transformed using an activation function, $f$, generating the output signal, $\hat{y} = f(z(x,w))$.  Activation functions are non-linear, allowing neural networks to characterize nonlinear relationships.  Activation functions are also differentiable allowing for efficient training through the backpropagation of model error during training. Single neurons are aggregated together to form a layer, and a network with more than one or more hidden layers is called a deep neural network (DNN).  Adding neurons and layers tends to increase the complexity of the problem the network is able to solve. More on deep neural networks can be found in \cite{goodfellow2016deep,chollet2021deep}. 

\subsection{Surrogate Modeling}\label{app:surrogate_modeling}

\begin{figure}
    \centering
    \includegraphics[width=0.7\textwidth]{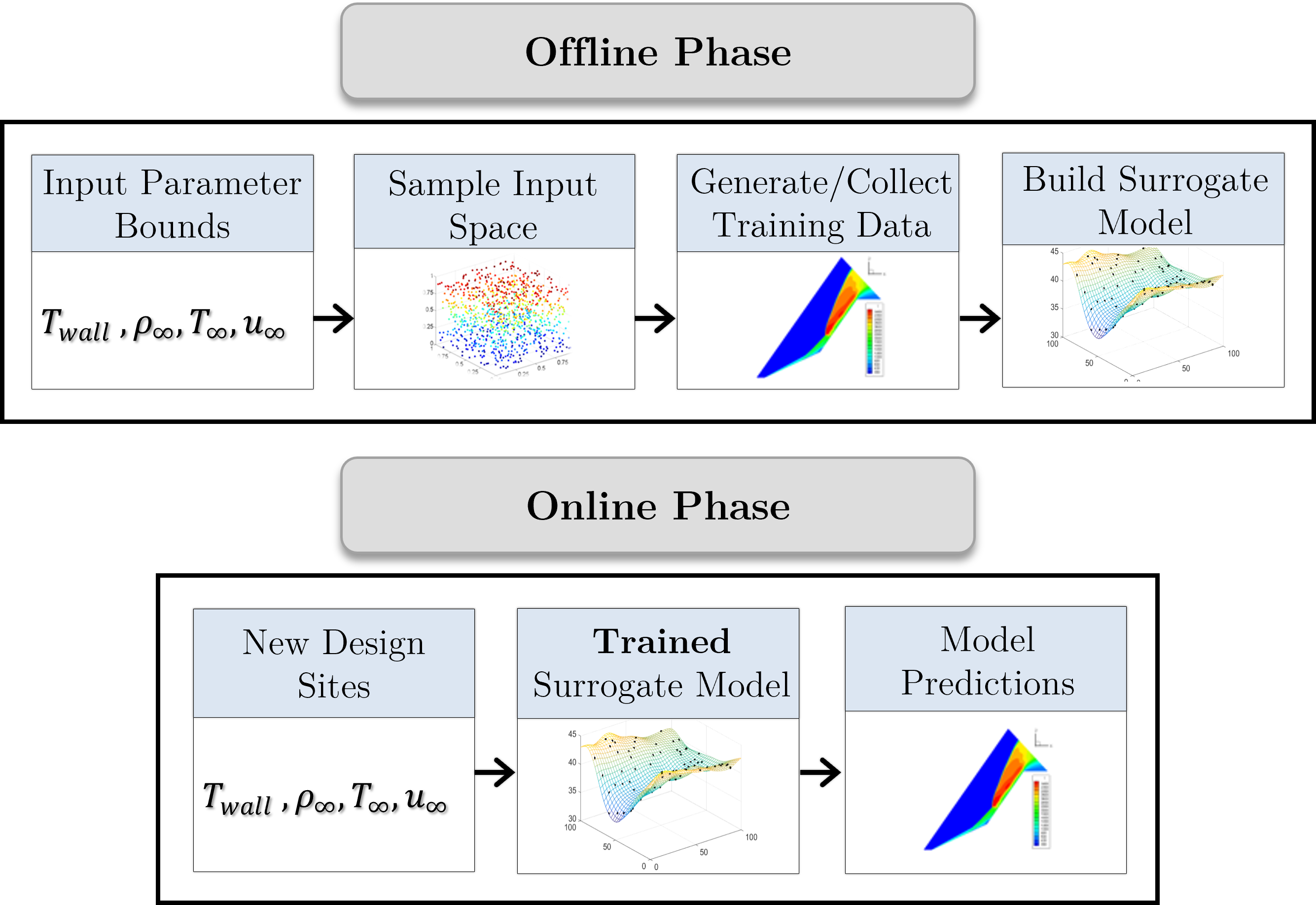}
    \caption{Surrogate modeling workflow}
    \label{fig:surrogatemodelingworkflow}
\end{figure}

A general surrogate modeling workflow is presented in this section and depicted in Fig.~\ref{fig:surrogatemodelingworkflow}. 

\[
\begin{array}{rlcll}
\makebox[2.5cm][l]{1) \text{Generate data}} & | & F(X_{\text{train}}) & = & Y_{\text{train}} \\
\makebox[2.5cm][l]{2) \text{Train model}} & | & F^*(X_{\text{train}}) & \mapsto & Y_{\text{train}} \\
\makebox[2.5cm][l]{3) \text{Evaluate model}} & | & F^*(X_{\text{DS}}) & \approx & Y_{\text{DS}}
\end{array}
\]

\subsection{N-Step Multi-fidelity Modeling}

\begin{figure}
    \centering
    \includegraphics[width=0.8\textwidth]{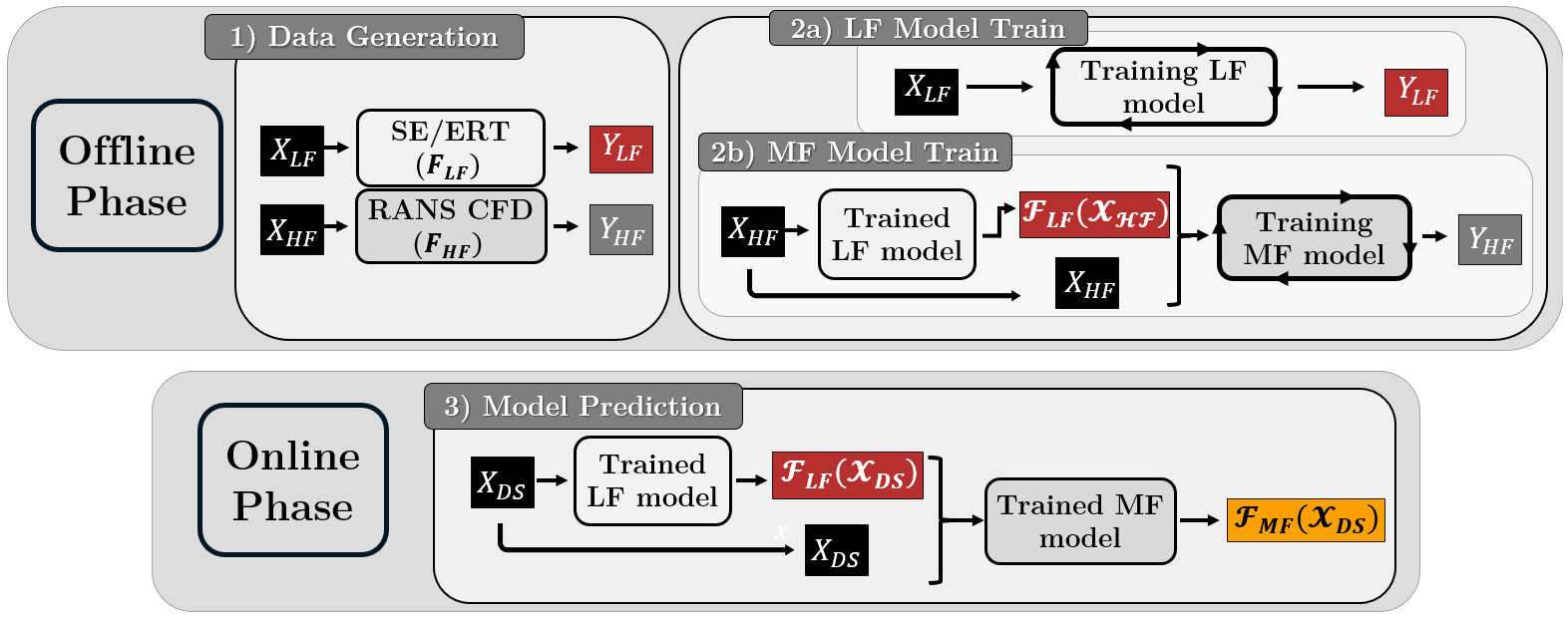}
    \caption{Multi-fidelity surrogate modeling workflow example, from \cite{korenyi2024josr}}
    \label{fig:MFsurrogatemodelingworkflow_example}
\end{figure}

Extending the idea in the above section (Sec.~\ref{app:surrogate_modeling}) to incorporate multiple data sources of varying fidelity. This algorithm follows the work by Guo et al. 2022 \cite{GUO2022114378} and implicitly, \cite{meng2020composite,motamed2020multi}. A more detailed discussion about multi-fidelity modeling can also be found in prior work~\cite{korenyi2023thesis}. An example multi-fidelity modeling workflow for Dataset 2 (Sec.~\ref{subsubsec:dataset2}) is depicted in Fig.~\ref{fig:MFsurrogatemodelingworkflow_example}

\[
\begin{array}{rlcll}
\makebox[4.5cm][l]{Generate HF data} & | & F_{HF}(X_{HF}) & = & Y_{HF} \\
\makebox[4.5cm][l]{Generate LF data} & | & F_{LF}(X_{LF}) & = & Y_{LF} \\
\makebox[4.5cm][l]{Train LF model} & | & F_{LF}^*(X_{LF}) & \mapsto & Y_{LF} \\
\makebox[4.5cm][l]{Evaluate LF model at HF points} & | & F_{LF}^*(X_{HF}) & \approx & \{ F_{LF}^*(X_{HF}^i) : 1 \leq i \leq N_{HF} \} \\
\makebox[4.5cm][l]{Train MF model} & | & F_{MF}^*(F_{LF}^*(X_{HF}), X_{HF}) & \mapsto & \{ Y_{HF} : 1 \leq i \leq N_{HF} \} \\
\makebox[4.5cm][l]{Evaluate MF model} & | & F_{MF}^*(F_{LF}^*(X_{DS}), X_{DS}) & \approx & \{ Y_{DS} : 1 \leq i \leq N_{DS} \}
\end{array}
\]

\section{Surrogate Modeling Pipeline Instructions}\label{app:sm_pipeline}
The pipeline overview is provided below. 

\begin{lstlisting}
model_data_dict = load_and_format_data(input_data.csv, output_data.csv)
model_data_dict = preprocess_data_pipeline(model_data_dict)
optimal_hyperparameters = tune_model(model_data_dict, hyperparameter_bounds)
trained_model = build_and_train_model(X_train, y_train,optimal_hyperparameters)
model_data_dict =  postprocess_data_pipeline(model_data_dict, trained_model)
\end{lstlisting}

\subsection{Configure Environment}
VSCode is the author's preferred IDE, although any environment capable of running Python will suffice. The framework is implemented in Python 3.9. A GPU can accelerate DNN training but is not required. Key Python dependencies are scikit-learn, TensorFlow, Keras, pandas, and SciPy~\cite{pedregosa2011scikit, abadi2016tensorflow, chollet2015keras, Pandas_mckinney2010data, 2020SciPy-NMeth}. To begin, download the code from the author's GitHub repository or request it directly from the author. Environment requirements and dependencies are listed in the GitHub repository. The following four files are the minimum required unless specific pre-processing scripts are needed for data cleaning, depending on the user's simulation or experimental data setup.

\begin{itemize}
    \item sm\_modeling\_notebook.ipynb (example workflow)
    \item scientific\_utils.py (main library)
    \item modeling\_plot.py (plotting tools)
    \item requirements.txt (environment requirements for pip installer)
\end{itemize}

\subsection{Clean and Format Data}
Before importing data into the framework, post-process the QoI data of the simulation or experiment into the following format: \texttt{(samples,scalar\_values,coordinates)} or $(n,m,l)$. Discussed in more detail in Sec.~\ref{subsec:dataset}. Pseudocode for a CBAERO example is provided below: 

\begin{algorithm}[H]
\caption{Processing CBAERO Output File}\label{alg:process_hgv_data}
\begin{algorithmic}[1]
\Procedure{ProcessData}{}
    \State Load the \texttt{.mat} data file    
    \State Extract input and output variable names from data using regular expressions
    \State Collect input parameter values from data structure
    \State Preallocate input and output parameters arrays
    \For{each $\beta$ value}
        \For{each Mach value}
            \For{each Altitude value}
                \For{each $\alpha$ value}
                    \State Store input parameter values in input array
                    \State Initialize temporary output values array
                    \State Extract and store output values
                \EndFor
            \EndFor
        \EndFor
    \EndFor
    \State Save collected data to \texttt{.mat} file
\EndProcedure
\end{algorithmic}
\end{algorithm}

Included in the GitHub (or upon request) are scripts that will convert CBAERO or US3D data into the required format. It is also helpful, if the size of the data permits, to save the data into .pkl files or .csv. This practice facilitates data sharing with other engineers working on the same datasets or researchers who wish to verify the work. The presented modeling framework has an example script for turning a NumPy 3D array into .txt files that can be ingested by the modeling pipeline, which helps when transferring data to air-gapped systems. 

\begin{figure}[h!]
    \centering
    \includegraphics[width=0.9\textwidth]{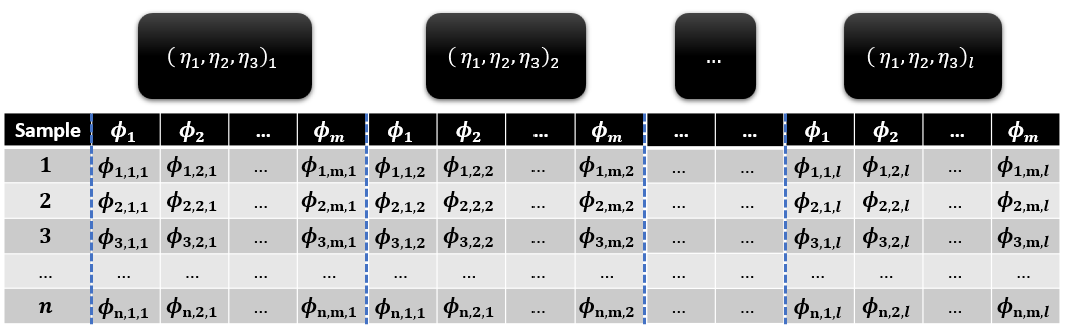}
    \caption{Data standard at training and inference for data-driven methods. 2D tensor, shape $(n, m*l)$.}
    \label{fig:data_standard_stacked}
\end{figure}

Initially, the data is imported in the format shown in Fig.\ref{fig:data_standard_stacked}. However, for the standard GPR and DNN models used in this framework, the data is rearranged as shown in Fig.\ref{fig:data_standard_stacked}. During training and inference, the data is processed as a 2D tensor, with all fluid scalars for each case stacked into a single vector. This approach has limitations. By disregarding the spatial relationships in the data, we lose the ability to utilize known geometrical or physical properties of the problem. This limitation is further explored in Sec.~\ref{sec:future_work}.

\begin{figure}[h!]
    \centering
    \includegraphics[width=0.9\textwidth]{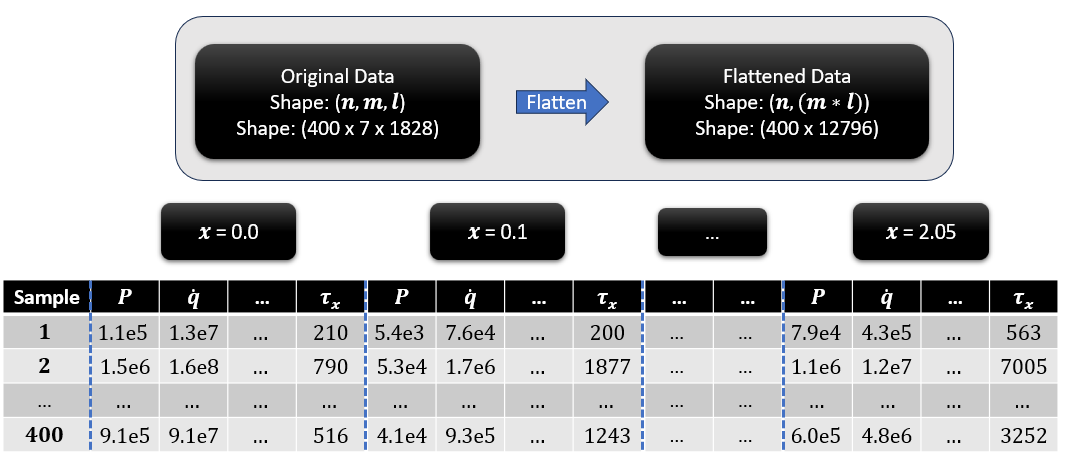}
    \caption{Data standard at training and inference example for Dataset 2. Shape (400, 12796)}
    \label{fig:data_standard_stacked_example}
\end{figure}

After cleaning and importing data, it is useful to conduct a preliminary data analysis. First, plot the input space using the included code or seaborn's~\cite{Waskom2021} \texttt{sns.pairplot(input\_data, corner=True, diag\_kind='kde')}. This helps determine if a statistically sound sampling strategy was used (particularly useful when using data provided by a third party) and assess parameter space coverage.

\subsection{Pre-process Data}
Before training, the data must be split into three cross-validation bins (training data, test data, and validation data) and normalized to have a mean of 0 and unit variance. The rationale for this preprocessing step is discussed in detail in prior work~\cite{korenyi2023thesis}. This framework automates the entire preprocessing pipeline, returning the processed data into the dictionary passed as an argument. For instance, \texttt{model\_data\_dict\_FL1 = preprocess\_data\_pipeline(model\_data\_dict, seed = settings.random\_state)}. The function first splits the data using scikit-learn's \texttt{split\_data\_cv} and then scales and stores the data using scikit-learn's \texttt{standard\_scaler}. The framework's modular design allows users to substitute \texttt{standard\_scaler} with any other scikit-learn scaler object if desired. Finally, the function consolidates the scaled and split data into NumPy arrays and stores the \texttt{standard\_scaler} objects for later use in inverse transformation after model predictions.

\subsection{Tune Model}
A hyperparameter is a user-defined value that controls model training and significantly influences model performance~\cite{feurer2019hyperparameter}. GPR-based models typically have few hyperparameters, such as length scale, output variance, and noise level~\cite{duvenaud_2014}. These hyperparameters are collectively represented as $\theta$, and their type and number depend on the kernel selection. Neural networks have significantly more hyperparameters, including network topology, size, learning rate, optimizer, and activation functions~\cite{chollet2021deep}.

The GPR model in this work optimizes hyperparameters during training using log-marginal likelihood (LMLL) to update kernel parameters. Selecting the appropriate kernel is the most important choice. The framework defaults to considering two primary kernels: Matern and RBF. Searching over these two kernel types, with broad hyperparameter bounds, can be effective, especially if the initial hyperparameters are within the correct order of magnitude before training and the optimizer is restarted a couple of times (typically n=2 or n=3). This not the most mathematically rigorous approach, but this method has been shown to work well on various datasets. Despite the ``default" kernel selections, the framework's modularity allows tuning over different kernels and kernel combinations.

Hyperparameter tuning for neural networks is more complex than for GPR models. This study uses a combination of KerasTuner~\cite{omalley2019kerastuner} and iterative comparison. KerasTuner, a hyperparameter optimization framework, reduces the workload associated with hyperparameter search but introduces secondary hyperparameters, such as selecting which hyperparameters to tune and setting bounds for the search space. These secondary hyperparameters are generally easier to manage than the primary model hyperparameters~\cite{goodfellow2016deep}. However, KerasTuner does not optimize for model sparsity and data compression while maximizing accuracy, often returning the upper bound of model size when layer width and depth are tuneable hyperparameters. Thus, KerasTuner was not exclusively used for tuning neural network models. One of the most crucial hyperparameters for a shallow DNN surrogate models are the number of layers and their width, which can be easily tuned with a vectorized \texttt{for} loop. 

A key aspect of tuning both exact GPR models and shallow DNNs is model convergence~\cite{korenyi2024josr}, similar to CFD grid convergence. The goal of model convergence is to use the smallest possible dataset for two reasons: 1) practical CFD data is large ($\mathcal{O}$(TBs)), and GPR models grow rapidly with the number of training points. Thus, small models offer good compression of the original dataset, which is desirable; and 2) model parsimony—using the smallest model and the optimal number of training points enhances model generalizability by reducing the risk of overfitting.

Once the model is tuned, the selected hyperparameters can be used in the next step (training), or the saved model can be kept for test and evaluation, thereby skipping the subsequent training step.

\subsection{Train Model}
The framework provides a function to build and train models using the specified hyperparameters. Currently, it supports two model types: Gaussian Process Regression (GPR) with scikit-learn and Deep Neural Networks (DNN) with Keras. The modular design of the framework enables the straightforward integration of other Python-based modeling techniques by subclassing.

\begin{lstlisting}
def build_and_train_model(X_train, y_train, model_type, path, project_name, hyperparam_dict):
    validated_hyperparams = validate_hyperparams(hyperparam_dict)
    model = build_model(model_type, validated_hyperparams)
    model_info = execute_training(model, X_train, y_train, validated_hyperparams)
    save_path, model_name = model_info['model'].save_model(fidelity_level=fidelity_level, path=path, project_name=project_name)
    metadata = model_info.items()
    save_versioned_pickle(filename, metadata, save_path)
    return model_info
\end{lstlisting}

\subsection{Analyze Model Results}
Evaluating the model using held-out test data is essential for understanding its generalization capability to new data within the original input parameter space. The \texttt{modeling\_plot.py} script features a function, \texttt{plot\_one\_to\_one\_comp()}, which creates a one-to-one scatter plot of the test data. This plot also provides summary statistics and details about the model's hyperparameters. Although not comprehensive, it delivers a clear snapshot of model performance. An $R^2$ value greater than 0.9 is considered indicative of good model performance~\cite{forrester2008engineering}.

\subsection{Post-Process Data}
After evaluating the model and generating predictions, the data must be restored to its original scale and shape, which was initially scaled to have a mean of zero and a variance of one. The function described below performs this task by first evaluating the model to generate predictions and then applying an inverse transformation to the data. The inverse-transformed model predictions are then stored in the model data dictionary.

\begin{lstlisting}
def postprocess_data_pipeline(model_data_dict: Dict, trained_model: Any) -> Dict:

    # Step 1: Generate inverse transformed predictions
    model_data_dict = generate_inverse_transformed_predictions(model_data_dict,trained_model)
    # Step 2: Verify the inverse transformation process
    verify_inverse_transformation_process(model_data_dict, generate_inverse_transformed_predictions)
    return model_data_dict
\end{lstlisting}

\subsection{Evaluate Model at New Design Sites}
Once the model is trained and validated, it is ready for online use. It can generate predictions for new input parameters by following a similar process. This involves evaluating the model at new design sites.

\begin{lstlisting}
def gen_new_pred_from_trained_model_pipeline(X_DS, hf_data_dict, lf_model_dict, mf_model_dict,write_to_csv) -> Dict:

    # Step 1: Make a new data dict at provided design sites (DS)
    new_DS_dict = create_DS_dict(hf_data_dict, X_DS)
    # Step 2: Update the new data dict
    new_DS_dict = add_vars_by_name_to_dict(model_data_dict=new_DS_dict)
    # Step 3: Scale the inputs using pre-fit scaler objects (fit during training)
    new_DS_dict = scale_and_store_new_DS_input(new_DS_dict, verbose=verbose)
    # Step 4: Generate the low-fidelity contributions to the multi-fidelity model input
    new_DS_dict = gen_data_for_multi_fidelity( lower_fidelity_model_dict, higher_fidelity_data_dict)
    # Step 5: Generate the multi-fidelity model predictions at new design sites. 
    new_DS_dict = generate_new_DS_inverse_transformed_predictions(model_data_dict, trained_model)
    if write_to_csv:
        export_DS_model_pred_to_CSV(model_data_dict)
    return new_DS_dict
\end{lstlisting}

\bibliography{sample}

\begin{thebibliography}{85}
\newcommand{\enquote}[1]{``#1''}
\providecommand{\natexlab}[1]{#1}
\providecommand{\url}[1]{\texttt{#1}}
\providecommand{\urlprefix}{URL }
\expandafter\ifx\csname urlstyle\endcsname\relax
  \providecommand{\doi}[1]{\discretionary{}{}{}https://doi.org/#1}\else
  \providecommand{\doi}[1]{\discretionary{}{}{}\urlstyle{rm}\url{https://doi.org/#1}}\fi

\bibitem[{Austin(2022)}]{austin_2022}
Austin, L., \enquote{2022 National Defense Strategy, Nuclear Posture Review, and Missile Defense Review,} , 2022.
\newblock \urlprefix\url{https://media.defense.gov/2022/Oct/27/2003103845/-1/-1/1/2022-NATIONAL-DEFENSE-STRATEGY-NPR-MDR.PDF}.

\bibitem[{Sayler(2022)}]{sayler_2022}
Sayler, K., \enquote{Hypersonic Weapons: Background and Issues For Congress,} , Mar 2022.
\newblock \urlprefix\url{https://crsreports.congress.gov/product/pdf/R/R45811}.

\bibitem[{Schmisseur(2015)}]{schmisseur2015hypersonics}
Schmisseur, J.~D., \enquote{Hypersonics into the 21st century: A perspective on AFOSR-sponsored research in aerothermodynamics,} \emph{Progress in Aerospace Sciences}, Vol.~72, 2015, pp. 3--16.

\bibitem[{Capaccio(2021)}]{capaccio_2021}
Capaccio, A., \enquote{Hypersonic Sticker Shock: U.S. Weapons May Run \$106 Million Each,} , Nov 2021.
\newblock \urlprefix\url{https://www.bloomberg.com/news/articles/2021-11-12/hypersonic-sticker-shock-u-s-weapons-may-run-106-million-each}.

\bibitem[{fy2(2021)}]{fy2022programacquisitioncostsbyweaponsystem_2021}
\enquote{Major Weapon Systems - U.S. Department of Defense,} , 2021.
\newblock \urlprefix\url{https://comptroller.defense.gov/Portals/45/Documents/defbudget/FY2022/FY2022_Weapons.pdf}.

\bibitem[{Ludwigson(2021)}]{GAO__hypersonics_2021}
Ludwigson, J., \enquote{Hypersonic Weapons,} , Mar 2021.
\newblock \urlprefix\url{https://www.gao.gov/assets/gao-21-378.pdf}.

\bibitem[{Marren and Lu(2002)}]{lu_2002}
Marren, D., and Lu, F., \emph{Advanced Hypersonic Test Facilities}, American Institute of Aeronautics and Astronautics, Reston ,VA, 2002.
\newblock \doi{10.2514/4.866678}, \urlprefix\url{https://arc.aiaa.org/doi/abs/10.2514/4.866678}.

\bibitem[{Hall et~al.(2000)Hall, Thomas, and Dowell}]{hall2000proper}
Hall, K.~C., Thomas, J.~P., and Dowell, E.~H., \enquote{Proper Orthogonal Decomposition Technique for Transonic Unsteady Aerodynamic Flows,} \emph{AIAA Journal}, Vol.~38, No.~10, 2000, pp. 1853--1862.
\newblock \doi{10.2514/2.867}, \urlprefix\url{https://doi.org/10.2514/2.867}.

\bibitem[{Venturi and Karniadakis(2004)}]{venturi2004gappy}
Venturi, D., and Karniadakis, G.~E., \enquote{Gappy data and reconstruction procedures for flow past a cylinder,} \emph{Journal of Fluid Mechanics}, Vol. 519, 2004, pp. 315--336.

\bibitem[{Lucia et~al.(2005)Lucia, Beran, and Silva}]{lucia2005aeroelastic}
Lucia, D.~J., Beran, P.~S., and Silva, W.~A., \enquote{Aeroelastic System Development Using Proper Orthogonal Decomposition and Volterra Theory,} \emph{Journal of Aircraft}, Vol.~42, No.~2, 2005, pp. 509--518.
\newblock \doi{10.2514/1.2176}, \urlprefix\url{https://doi.org/10.2514/1.2176}.

\bibitem[{Skujins and Cesnik(2014)}]{skujins2014reduced}
Skujins, T., and Cesnik, C. E.~S., \enquote{Reduced-Order Modeling of Unsteady Aerodynamics Across Multiple Mach Regimes,} \emph{Journal of Aircraft}, Vol.~51, No.~6, 2014, pp. 1681--1704.
\newblock \doi{10.2514/1.C032222}, \urlprefix\url{https://doi.org/10.2514/1.C032222}.

\bibitem[{Rowley and Dawson(2017)}]{rowley2017model}
Rowley, C.~W., and Dawson, S.~T., \enquote{Model Reduction for Flow Analysis and Control,} \emph{Annual Review of Fluid Mechanics}, Vol.~49, No.~1, 2017, pp. 387--417.
\newblock \doi{10.1146/annurev-fluid-010816-060042}, \urlprefix\url{https://doi.org/10.1146/annurev-fluid-010816-060042}.

\bibitem[{Kutz(2017)}]{kutz2017deep}
Kutz, J.~N., \enquote{Deep learning in fluid dynamics,} \emph{Journal of Fluid Mechanics}, Vol. 814, 2017, p. 1–4.
\newblock \doi{10.1017/jfm.2016.803}.

\bibitem[{Duraisamy et~al.(2019)Duraisamy, Iaccarino, and Xiao}]{duraisamy2019turbulence}
Duraisamy, K., Iaccarino, G., and Xiao, H., \enquote{Turbulence Modeling in the Age of Data,} \emph{Annual Review of Fluid Mechanics}, Vol.~51, No.~1, 2019, pp. 357--377.
\newblock \doi{10.1146/annurev-fluid-010518-040547}, \urlprefix\url{https://doi.org/10.1146/annurev-fluid-010518-040547}.

\bibitem[{Brunton et~al.(2020)Brunton, Noack, and Koumoutsakos}]{brunton2020machine}
Brunton, S.~L., Noack, B.~R., and Koumoutsakos, P., \enquote{Machine Learning for Fluid Mechanics,} \emph{Annual Review of Fluid Mechanics}, Vol.~52, No.~1, 2020, pp. 477--508.
\newblock \doi{10.1146/annurev-fluid-010719-060214}, \urlprefix\url{https://doi.org/10.1146/annurev-fluid-010719-060214}.

\bibitem[{Dreyer et~al.(2021)Dreyer, Grier, McNamara, and Orr}]{Dreyer2021}
Dreyer, E.~R., Grier, B.~J., McNamara, J.~J., and Orr, B.~C., \enquote{Rapid Steady-State Hypersonic Aerothermodynamic Loads Prediction Using Reduced Fidelity Models,} \emph{Journal of Aircraft}, Vol.~58, No.~3, 2021, pp. 663--676.
\newblock \doi{10.2514/1.C035969}, \urlprefix\url{https://doi.org/10.2514/1.C035969}.

\bibitem[{Barnett et~al.(2023)Barnett, Farhat, and Maday}]{barnett2023neural}
Barnett, J., Farhat, C., and Maday, Y., \enquote{Neural-network-augmented projection-based model order reduction for mitigating the Kolmogorov barrier to reducibility,} \emph{Journal of Computational Physics}, Vol. 492, 2023, p. 112420.

\bibitem[{Needels and Alonso(2023)}]{needels2023efficient}
Needels, J.~T., and Alonso, J.~J., \enquote{Efficient Global Optimization for Multidisciplinary Conceptual Design of Hypersonic Vehicles,} \emph{AIAA Aviation 2023 Forum}, 2023, p. 3718.

\bibitem[{Shukla et~al.(2024)Shukla, Oommen, Peyvan, Penwarden, Plewacki, Bravo, Ghoshal, Kirby, and Karniadakis}]{shukla_deep_2024}
Shukla, K., Oommen, V., Peyvan, A., Penwarden, M., Plewacki, N., Bravo, L., Ghoshal, A., Kirby, R.~M., and Karniadakis, G.~E., \enquote{Deep neural operators as accurate surrogates for shape optimization,} \emph{Engineering Applications of Artificial Intelligence}, Vol. 129, 2024, p. 107615.
\newblock \doi{10.1016/j.engappai.2023.107615}, \urlprefix\url{https://linkinghub.elsevier.com/retrieve/pii/S0952197623017992}.

\bibitem[{Forrester et~al.(2008)Forrester, Sobester, and Keane}]{forrester2008engineering}
Forrester, A., Sobester, A., and Keane, A., \emph{Engineering Design via Surrogate Modelling: A Practical Guide}, John Wiley \& Sons, 2008.
\newblock \doi{10.1002/9780470770801}.

\bibitem[{Forrester et~al.(2007)Forrester, Sóbester, and Keane}]{forrester2007multi}
Forrester, A.~I., Sóbester, A., and Keane, A.~J., \enquote{Multi-fidelity optimization via surrogate modelling,} \emph{Proceedings of the Royal Society A: Mathematical, Physical and Engineering Sciences}, Vol. 463, No. 2088, 2007, pp. 3251--3269.
\newblock \doi{10.1098/rspa.2007.1900}, \urlprefix\url{https://royalsocietypublishing.org/doi/abs/10.1098/rspa.2007.1900}.

\bibitem[{Raissi et~al.(2017)Raissi, Perdikaris, and Karniadakis}]{raissi2017inferring}
Raissi, M., Perdikaris, P., and Karniadakis, G.~E., \enquote{Inferring solutions of differential equations using noisy multi-fidelity data,} \emph{Journal of Computational Physics}, Vol. 335, 2017, pp. 736--746.

\bibitem[{Meng and Karniadakis(2020)}]{meng2020composite}
Meng, X., and Karniadakis, G.~E., \enquote{A composite neural network that learns from multi-fidelity data: Application to function approximation and inverse PDE problems,} \emph{Journal of Computational Physics}, Vol. 401, 2020, p. 109020.
\newblock \doi{10.1016/j.jcp.2019.109020}.

\bibitem[{Karniadakis et~al.(2021)Karniadakis, Kevrekidis, Lu, Perdikaris, Wang, and Yang}]{karniadakis2021physics}
Karniadakis, G.~E., Kevrekidis, I.~G., Lu, L., Perdikaris, P., Wang, S., and Yang, L., \enquote{Physics-informed machine learning,} \emph{Nature Reviews Physics}, Vol.~3, No.~6, 2021, pp. 422--440.

\bibitem[{Pestourie et~al.(2023)Pestourie, Mroueh, Rackauckas, Das, and Johnson}]{pestourie_physics-enhanced_2023}
Pestourie, R., Mroueh, Y., Rackauckas, C., Das, P., and Johnson, S.~G., \enquote{Physics-enhanced deep surrogates for partial differential equations,} \emph{Nature Machine Intelligence}, Vol.~5, No.~12, 2023, pp. 1458--1465.
\newblock \doi{10.1038/s42256-023-00761-y}, \urlprefix\url{https://www.nature.com/articles/s42256-023-00761-y}.

\bibitem[{Raissi et~al.(2019)Raissi, Perdikaris, and Karniadakis}]{raissi2019physics}
Raissi, M., Perdikaris, P., and Karniadakis, G.~E., \enquote{Physics-informed neural networks: A deep learning framework for solving forward and inverse problems involving nonlinear partial differential equations,} \emph{Journal of Computational physics}, Vol. 378, 2019, pp. 686--707.

\bibitem[{Krishnapriyan et~al.(2021)Krishnapriyan, Gholami, Zhe, Kirby, and Mahoney}]{krishnapriyan2021characterizing}
Krishnapriyan, A., Gholami, A., Zhe, S., Kirby, R., and Mahoney, M.~W., \enquote{Characterizing possible failure modes in physics-informed neural networks,} \emph{Advances in Neural Information Processing Systems}, Vol.~34, 2021, pp. 26548--26560.

\bibitem[{Lu et~al.(2021{\natexlab{a}})Lu, Jin, and Karniadakis}]{lu_deeponet_2021}
Lu, L., Jin, P., and Karniadakis, G.~E., \enquote{{DeepONet}: {Learning} nonlinear operators for identifying differential equations based on the universal approximation theorem of operators,} \emph{Nature Machine Intelligence}, Vol.~3, No.~3, 2021{\natexlab{a}}, pp. 218--229.
\newblock \doi{10.1038/s42256-021-00302-5}, \urlprefix\url{http://arxiv.org/abs/1910.03193}, arXiv:1910.03193 [cs, stat].

\bibitem[{Kovachki et~al.(2023)Kovachki, Li, Liu, Azizzadenesheli, Bhattacharya, Stuart, and Anandkumar}]{kovachki2023neural}
Kovachki, N., Li, Z., Liu, B., Azizzadenesheli, K., Bhattacharya, K., Stuart, A., and Anandkumar, A., \enquote{Neural operator: Learning maps between function spaces with applications to pdes,} \emph{Journal of Machine Learning Research}, Vol.~24, No.~89, 2023, pp. 1--97.

\bibitem[{Azizzadenesheli et~al.(2024)Azizzadenesheli, Kovachki, Li, Liu-Schiaffini, Kossaifi, and Anandkumar}]{azizzadenesheli2024neural}
Azizzadenesheli, K., Kovachki, N., Li, Z., Liu-Schiaffini, M., Kossaifi, J., and Anandkumar, A., \enquote{Neural operators for accelerating scientific simulations and design,} \emph{Nature Reviews Physics}, 2024, pp. 1--9.

\bibitem[{Chen and Chen(1995)}]{chen1995universal}
Chen, T., and Chen, H., \enquote{Universal approximation to nonlinear operators by neural networks with arbitrary activation functions and its application to dynamical systems,} \emph{IEEE transactions on neural networks}, Vol.~6, No.~4, 1995, pp. 911--917.

\bibitem[{Li et~al.(2020)Li, Kovachki, Azizzadenesheli, Liu, Bhattacharya, Stuart, and Anandkumar}]{li2020fourier}
Li, Z., Kovachki, N., Azizzadenesheli, K., Liu, B., Bhattacharya, K., Stuart, A., and Anandkumar, A., \enquote{Fourier neural operator for parametric partial differential equations,} \emph{arXiv preprint arXiv:2010.08895}, 2020.

\bibitem[{Lu et~al.(2022)Lu, Meng, Cai, Mao, Goswami, Zhang, and Karniadakis}]{lu2022comprehensive}
Lu, L., Meng, X., Cai, S., Mao, Z., Goswami, S., Zhang, Z., and Karniadakis, G.~E., \enquote{A comprehensive and fair comparison of two neural operators (with practical extensions) based on fair data,} \emph{Computer Methods in Applied Mechanics and Engineering}, Vol. 393, 2022, p. 114778.

\bibitem[{Li et~al.(2021)Li, Zheng, Kovachki, Jin, Chen, Liu, Azizzadenesheli, and Anandkumar}]{li2021physics}
Li, Z., Zheng, H., Kovachki, N., Jin, D., Chen, H., Liu, B., Azizzadenesheli, K., and Anandkumar, A., \enquote{Physics-informed neural operator for learning partial differential equations,} \emph{ACM/JMS Journal of Data Science}, 2021.

\bibitem[{Goswami et~al.(2023)Goswami, Bora, Yu, and Karniadakis}]{goswami2023physics}
Goswami, S., Bora, A., Yu, Y., and Karniadakis, G.~E., \enquote{Physics-informed deep neural operator networks,} \emph{Machine Learning in Modeling and Simulation: Methods and Applications}, Springer, 2023, pp. 219--254.

\bibitem[{Di~Leoni et~al.(2023)Di~Leoni, Lu, Meneveau, Karniadakis, and Zaki}]{di2023neural}
Di~Leoni, P.~C., Lu, L., Meneveau, C., Karniadakis, G.~E., and Zaki, T.~A., \enquote{Neural operator prediction of linear instability waves in high-speed boundary layers,} \emph{Journal of Computational Physics}, Vol. 474, 2023, p. 111793.

\bibitem[{Pedregosa et~al.(2011)Pedregosa, Varoquaux, Gramfort, Michel, Thirion, Grisel, Blondel, Prettenhofer, Weiss, Dubourg, Vanderplas, Passos, Cournapeau, Brucher, Perrot, and Duchesnay}]{pedregosa2011scikit}
Pedregosa, F., Varoquaux, G., Gramfort, A., Michel, V., Thirion, B., Grisel, O., Blondel, M., Prettenhofer, P., Weiss, R., Dubourg, V., Vanderplas, J., Passos, A., Cournapeau, D., Brucher, M., Perrot, M., and Duchesnay, E., \enquote{Scikit-learn: Machine Learning in {P}ython,} \emph{Journal of Machine Learning Research}, Vol.~12, 2011, pp. 2825--2830.

\bibitem[{Abadi et~al.(2016)Abadi, Barham, Chen, Chen, Davis, Dean, Devin, Ghemawat, Irving, Isard et~al.}]{abadi2016tensorflow}
Abadi, M., Barham, P., Chen, J., Chen, Z., Davis, A., Dean, J., Devin, M., Ghemawat, S., Irving, G., Isard, M., et~al., \enquote{Tensorflow: A system for large-scale machine learning,} \emph{12th USENIX Symposium on Operating Systems Design and Implementation (OSDI 16)}, 2016, pp. 265--283.

\bibitem[{Chollet et~al.(2015)}]{chollet2015keras}
Chollet, F., et~al., \enquote{Keras,} , 2015.
\newblock \urlprefix\url{https://github.com/fchollet/keras}.

\bibitem[{Lophaven et~al.(2002)Lophaven, Nielsen, and Sondergaard}]{lophaven2002matlab}
Lophaven, S.~N., Nielsen, H.~B., and Sondergaard, J., \enquote{DACE - A MATLAB Kriging Toolbox,} \emph{Technical University of Denmark, Kongens Lyngby, Technical Report No. IMM-TR-2002-12}, 2002.

\bibitem[{Bouhlel et~al.(2019)Bouhlel, Hwang, Bartoli, Lafage, Morlier, and Martins}]{SMT2019}
Bouhlel, M.~A., Hwang, J.~T., Bartoli, N., Lafage, R., Morlier, J., and Martins, J. R. R.~A., \enquote{A Python surrogate modeling framework with derivatives,} \emph{Advances in Engineering Software}, 2019, p. 102662.
\newblock \doi{https://doi.org/10.1016/j.advengsoft.2019.03.005}.

\bibitem[{Saves et~al.(2024)Saves, Lafage, Bartoli, Diouane, Bussemaker, Lefebvre, Hwang, Morlier, and Martins}]{saves2024smt}
Saves, P., Lafage, R., Bartoli, N., Diouane, Y., Bussemaker, J., Lefebvre, T., Hwang, J.~T., Morlier, J., and Martins, J. R. R.~A., \enquote{{SMT 2.0: A} Surrogate Modeling Toolbox with a focus on Hierarchical and Mixed Variables Gaussian Processes,} \emph{Advances in Engineering Sofware}, Vol. 188, 2024, p. 103571.
\newblock \doi{https://doi.org/10.1016/j.advengsoft.2023.103571}.

\bibitem[{Korenyi-Both et~al.(2024)Korenyi-Both, McNamara, Willems, and Reasor}]{korenyi2024josr}
Korenyi-Both, T., McNamara, J.~J., Willems, J.~A., and Reasor, D.~A., \enquote{Assessment of Multifidelity Surrogate Approaches for Expedient Loads Prediction in High-Speed Flows,} \emph{AIAA Journal of Spacecraft and Rockets}, 2024.
\newblock \doi{10.2514/1.A36043}, {A}rticle in Advance.

\bibitem[{Korenyi-Both(2023)}]{korenyi2023thesis}
Korenyi-Both, T.~E., \enquote{Assessment of Multi-Fidelity Surrogate Methods for Expedient Loads Prediction in High-Speed Flows,} Master's thesis, The Ohio State University, 2023.

\bibitem[{Korenyi-Both et~al.(2023)Korenyi-Both, McNamara, Willems, and Reasor}]{korenyi2023conference}
Korenyi-Both, T., McNamara, J.~J., Willems, J.~A., and Reasor, D.~A., \enquote{Assessment of Multi-Fidelity Surrogate Approaches for Expedient Loads Prediction in High-Speed Flows,} \emph{AIAA AVIATION 2023 Forum}, 2023, p. 3844.

\bibitem[{Sacks et~al.(1989)Sacks, Welch, Mitchell, and Wynn}]{Sack1989DoE}
Sacks, J., Welch, W.~J., Mitchell, T.~J., and Wynn, H.~P., \enquote{Design and Analysis of Computer Experiments,} \emph{Statistical Science}, Vol.~4, No.~4, 1989, pp. 409--423.
\newblock \urlprefix\url{http://www.jstor.org/stable/2245858}.

\bibitem[{Lu et~al.(2021{\natexlab{b}})Lu, Meng, Mao, and Karniadakis}]{lu2021deepxde}
Lu, L., Meng, X., Mao, Z., and Karniadakis, G.~E., \enquote{{DeepXDE}: A deep learning library for solving differential equations,} \emph{SIAM Review}, Vol.~63, No.~1, 2021{\natexlab{b}}, pp. 208--228.
\newblock \doi{10.1137/19M1274067}.

\bibitem[{Dreyer et~al.(2017)Dreyer, Klock, Grier, McNamara, and Cesnik}]{dreyer2017multi}
Dreyer, E., Klock, R., Grier, B., McNamara, J.~J., and Cesnik, C.~E., \enquote{Multi-discipline modeling of complete hypersonic vehicles using cfd surrogates,} \emph{58th AIAA/ASCE/AHS/ASC Structures, Structural Dynamics, and Materials Conference}, 2017, p. 0182.

\bibitem[{Reasor et~al.(2021)Reasor, Baxley, Goble, Minsavage-Davis, and Willems~Jr.}]{reasor_2021}
Reasor, D., Baxley, J., Goble, J., Minsavage-Davis, K., and Willems~Jr., J., \enquote{Fast Running Decoders for CFD Reproduction Presentation At 2021 AFOSR Computational Math Program Review,} Air Force Office of Scientific Research, 2021.

\bibitem[{Dreyer(2021)}]{dreyer2021dissertation}
Dreyer, E.~R., \enquote{Assessment of Reduced Fidelity Modeling of a Maneuvering Hypersonic Vehicle,} Ph.D. thesis, The Ohio State University, 2021.

\bibitem[{Vanderwyst et~al.(2016)Vanderwyst, Shelton, and Martin}]{vanderwyst2016computationally}
Vanderwyst, A., Shelton, A., and Martin, C.~L., \enquote{A Computationally Efficient, Multi-Fidelity Assessment of Jet Interactions for Highly Maneuverable Missiles,} \emph{34th AIAA Applied Aerodynamics Conference}, 2016, p. 4333.
\newblock \doi{10.2514/6.2016-4333}, \urlprefix\url{https://arc.aiaa.org/doi/abs/10.2514/6.2016-4333}.

\bibitem[{Queipo et~al.(2005)Queipo, Haftka, Shyy, Goel, Vaidyanathan, and Tucker}]{queipo2005surrogate}
Queipo, N.~V., Haftka, R.~T., Shyy, W., Goel, T., Vaidyanathan, R., and Tucker, P.~K., \enquote{Surrogate-based analysis and optimization,} \emph{Progress in aerospace sciences}, Vol.~41, No.~1, 2005, pp. 1--28.

\bibitem[{Kinney(2004)}]{kinney2004aero}
Kinney, D., \enquote{Aero-thermodynamics for conceptual design,} \emph{42nd AIAA Aerospace Sciences Meeting and Exhibit}, 2004, p.~31.

\bibitem[{Candler et~al.(2015)Candler, Johnson, Nompelis, Gidzak, Subbareddy, and Barnhardt}]{candlerUS3D2015}
Candler, G.~V., Johnson, H.~B., Nompelis, I., Gidzak, V.~M., Subbareddy, P.~K., and Barnhardt, M., \enquote{Development of the US3D Code for Advanced Compressible and Reacting Flow Simulations,} \emph{53rd AIAA Aerospace Sciences Meeting}, AIAA, 2015, pp. 1--31.
\newblock \doi{10.2514/6.2015-1893}, \urlprefix\url{https://arc.aiaa.org/doi/abs/10.2514/6.2015-1893}.

\bibitem[{Anderson(2019)}]{anderson_2019}
Anderson, J.~D., \emph{Hypersonic and High-Temperature Gas Dynamics}, American Institute of Aeronautics and Astronautics, Inc., 2019.
\newblock \doi{10.2514/4.105142}.

\bibitem[{Eckert(1956)}]{Eckert_1956}
Eckert, E. R.~G., \enquote{{Engineering Relations for Heat Transfer and Friction in High-Velocity Laminar and Turbulent Boundary-Layer Flow Over Surfaces With Constant Pressure and Temperature},} \emph{Transactions of the American Society of Mechanical Engineers}, Vol.~78, No.~6, 1956, pp. 1273--1283.
\newblock \doi{10.1115/1.4014011}, \urlprefix\url{https://doi.org/10.1115/1.4014011}.

\bibitem[{Wadhams et~al.(2014)Wadhams, Holden, and MacLean}]{wadhams2014comparison}
Wadhams, T., Holden, M., and MacLean, M., \enquote{Comparison of experimental and computational results from blind turbulent shock wave interaction study over cone flare and hollow cylinder flare configuration,} \emph{AIAA Aviation. Atlanta, GA}, 2014.

\bibitem[{Guo et~al.(2022)Guo, Manzoni, Amendt, Conti, and Hesthaven}]{GUO2022114378}
Guo, M., Manzoni, A., Amendt, M., Conti, P., and Hesthaven, J.~S., \enquote{Multi-fidelity regression using artificial neural networks: Efficient approximation of parameter-dependent output quantities,} \emph{Computer Methods in Applied Mechanics and Engineering}, Vol. 389, 2022, p. 114378.
\newblock \doi{https://doi.org/10.1016/j.cma.2021.114378}, \urlprefix\url{https://www.sciencedirect.com/science/article/pii/S0045782521006411}.

\bibitem[{Motamed(2020)}]{motamed2020multi}
Motamed, M., \enquote{A Multi-Fidelity Neural Network Surrogate Sampling Method For Uncertainty Quantification,} \emph{International Journal for Uncertainty Quantification}, Vol.~10, No.~4, 2020.
\newblock \doi{10.1615/Int.J.UncertaintyQuantification.2020031957}.

\bibitem[{Van~Rossum and Drake(2009)}]{python}
Van~Rossum, G., and Drake, F.~L., \emph{Python 3 Reference Manual}, CreateSpace, Scotts Valley, CA, 2009.

\bibitem[{{W}es {M}c{K}inney(2010)}]{Pandas_mckinney2010data}
{W}es {M}c{K}inney, \enquote{{D}ata {S}tructures for {S}tatistical {C}omputing in {P}ython,} \emph{{P}roceedings of the 9th {P}ython in {S}cience {C}onference}, edited by {S}t\'efan van~der {W}alt and {J}arrod {M}illman, 2010, pp. 56 -- 61.
\newblock \doi{10.25080/Majora-92bf1922-00a}.

\bibitem[{Harris et~al.(2020)Harris, Millman, van~der Walt, Gommers, Virtanen, Cournapeau, Wieser, Taylor, Berg, Smith, Kern, Picus, Hoyer, van Kerkwijk, Brett, Haldane, del R{\'{i}}o, Wiebe, Peterson, G{\'{e}}rard-Marchant, Sheppard, Reddy, Weckesser, Abbasi, Gohlke, and Oliphant}]{harris2020array}
Harris, C.~R., Millman, K.~J., van~der Walt, S.~J., Gommers, R., Virtanen, P., Cournapeau, D., Wieser, E., Taylor, J., Berg, S., Smith, N.~J., Kern, R., Picus, M., Hoyer, S., van Kerkwijk, M.~H., Brett, M., Haldane, A., del R{\'{i}}o, J.~F., Wiebe, M., Peterson, P., G{\'{e}}rard-Marchant, P., Sheppard, K., Reddy, T., Weckesser, W., Abbasi, H., Gohlke, C., and Oliphant, T.~E., \enquote{Array programming with {NumPy},} \emph{Nature}, Vol. 585, No. 7825, 2020, pp. 357--362.
\newblock \doi{10.1038/s41586-020-2649-2}, \urlprefix\url{https://doi.org/10.1038/s41586-020-2649-2}.

\bibitem[{Gal and Ghahramani(2016)}]{pmlr-v48-gal16}
Gal, Y., and Ghahramani, Z., \enquote{Dropout as a Bayesian Approximation: Representing Model Uncertainty in Deep Learning,} \emph{Proceedings of The 33rd International Conference on Machine Learning}, Proceedings of Machine Learning Research, Vol.~48, edited by M.~F. Balcan and K.~Q. Weinberger, PMLR, New York, New York, USA, 2016, pp. 1050--1059.
\newblock \urlprefix\url{https://proceedings.mlr.press/v48/gal16.html}.

\bibitem[{Abdar et~al.(2021)Abdar, Pourpanah, Hussain, Rezazadegan, Liu, Ghavamzadeh, Fieguth, Cao, Khosravi, Acharya, Makarenkov, and Nahavandi}]{abdar2021review}
Abdar, M., Pourpanah, F., Hussain, S., Rezazadegan, D., Liu, L., Ghavamzadeh, M., Fieguth, P., Cao, X., Khosravi, A., Acharya, U.~R., Makarenkov, V., and Nahavandi, S., \enquote{A review of uncertainty quantification in deep learning: {Techniques}, applications and challenges,} \emph{Information Fusion}, Vol.~76, 2021, pp. 243--297.
\newblock \doi{https://doi.org/10.1016/j.inffus.2021.05.008}, \urlprefix\url{https://www.sciencedirect.com/science/article/pii/S1566253521001081}.

\bibitem[{Duvenaud(2014)}]{duvenaud_2014}
Duvenaud, D., \enquote{Automatic Model Construction with Gaussian Processes,} Ph.D. thesis, University of Cambridge, 2014.
\newblock \doi{10.17863/CAM.14087}.

\bibitem[{O'Malley et~al.(2019)O'Malley, Bursztein, Long, Chollet, Jin, Invernizzi et~al.}]{omalley2019kerastuner}
O'Malley, T., Bursztein, E., Long, J., Chollet, F., Jin, H., Invernizzi, L., et~al., \enquote{KerasTuner,} \url{https://github.com/keras-team/keras-tuner}, 2019.

\bibitem[{Bruinsma et~al.(2020)Bruinsma, Perim, Tebbutt, Hosking, Solin, and Turner}]{bruinsma_scalable_2020}
Bruinsma, W.~P., Perim, E., Tebbutt, W., Hosking, J.~S., Solin, A., and Turner, R.~E., \enquote{Scalable {Exact} {Inference} in {Multi}-{Output} {Gaussian} {Processes},} , Jul. 2020.
\newblock \urlprefix\url{http://arxiv.org/abs/1911.06287}, arXiv:1911.06287 [cs, stat].

\bibitem[{Gardner et~al.(2018)Gardner, Pleiss, Weinberger, Bindel, and Wilson}]{gardner2018gpytorch}
Gardner, J., Pleiss, G., Weinberger, K.~Q., Bindel, D., and Wilson, A.~G., \enquote{Gpytorch: Blackbox matrix-matrix gaussian process inference with gpu acceleration,} \emph{Advances in neural information processing systems}, Vol.~31, 2018.

\bibitem[{Falkiewicz et~al.(2011)Falkiewicz, Cesnik, Crowell, and McNamara}]{falkiewicz2011reduced}
Falkiewicz, N.~J., Cesnik, C.~E., Crowell, A.~R., and McNamara, J.~J., \enquote{Reduced-order aerothermoelastic framework for hypersonic vehicle control simulation,} \emph{AIAA journal}, Vol.~49, No.~8, 2011, pp. 1625--1646.

\bibitem[{Bostanabad et~al.(2019)Bostanabad, Chan, Wang, Zhu, and Chen}]{bostanabad_globally_2019}
Bostanabad, R., Chan, Y.-C., Wang, L., Zhu, P., and Chen, W., \enquote{Globally {Approximate} {Gaussian} {Processes} for {Big} {Data} {With} {Application} to {Data}-{Driven} {Metamaterials} {Design},} \emph{Journal of Mechanical Design}, Vol. 141, No.~11, 2019.
\newblock \doi{10.1115/1.4044257}, \urlprefix\url{https://doi.org/10.1115/1.4044257}.

\bibitem[{Bouhlel and Martins(2019)}]{bouhlel2019gradient}
Bouhlel, M.~A., and Martins, J.~R., \enquote{Gradient-enhanced kriging for high-dimensional problems,} \emph{Engineering with Computers}, Vol.~35, No.~1, 2019, pp. 157--173.

\bibitem[{Needels and Alonso(2024)}]{needels2024trajectory}
Needels, J.~T., and Alonso, J.~J., \enquote{Trajectory-informed Sampling for Efficient Construction of Multi-fidelity Surrogate Models for Hypersonic Vehicles,} \emph{AIAA SCITECH 2024 Forum}, 2024, p. 1013.

\bibitem[{Kamath(2022)}]{kamath2022intelligent}
Kamath, C., \enquote{Intelligent sampling for surrogate modeling, hyperparameter optimization, and data analysis,} \emph{Machine Learning with Applications}, Vol.~9, 2022, p. 100373.

\bibitem[{Krige(1951)}]{krige1951statistical}
Krige, D., \enquote{A statistical approach to some basic mine valuation problems on the Witwatersrand,} \emph{Journal of the Southern African Institute of Mining and Metallurgy}, Vol.~52, No.~6, 1951, pp. 119--139.
\newblock \doi{10.10520/AJA0038223X\_4792}, \urlprefix\url{https://journals.co.za/doi/abs/10.10520/AJA0038223X_4792}.

\bibitem[{Matheron(1963)}]{matheron1963principles}
Matheron, G., \enquote{{Principles of geostatistics},} \emph{Economic Geology}, Vol.~58, No.~8, 1963, pp. 1246--1266.
\newblock \doi{10.2113/gsecongeo.58.8.1246}, \urlprefix\url{https://doi.org/10.2113/gsecongeo.58.8.1246}.

\bibitem[{Wiener(1964)}]{wiener1949extrapolation}
Wiener, N., \emph{Extrapolation, Interpolation, and Smoothing of Stationary Time Series}, The MIT Press, 1964.

\bibitem[{Kolmogoroff(1941)}]{kolmogoroff1941interpolation}
Kolmogoroff, A., \enquote{Interpolation und Extrapolation von stationären zufälligen Folgen,} \emph{Izvestiya Rossiiskoi Akademii Nauk. Seriya Matematicheskaya}, Vol.~5, No.~1, 1941, pp. 3--14.

\bibitem[{Rasmussen and Williams(2005)}]{Rasmussen2006}
Rasmussen, C.~E., and Williams, C. K.~I., \emph{{Gaussian Processes for Machine Learning}}, The MIT Press, 2005.
\newblock \doi{10.7551/mitpress/3206.001.0001}, \urlprefix\url{https://doi.org/10.7551/mitpress/3206.001.0001}.

\bibitem[{McCulloch and Pitts(1943)}]{mcculloch_pitts_1943}
McCulloch, W.~S., and Pitts, W., \enquote{A logical calculus of the ideas immanent in nervous activity,} \emph{The Bulletin of Mathematical Biophysics}, Vol.~5, No.~4, 1943, p. 115–133.
\newblock \doi{10.1007/bf02478259}.

\bibitem[{Rosenblatt(1957)}]{rosenblatt_1957}
Rosenblatt, F., \enquote{The perceptron - A perceiving and recognizing automaton,} Tech. Rep. 85-460-1, Cornell Aeronautical Laboratory, Ithaca, New York, January 1957.

\bibitem[{Goodfellow et~al.(2016)Goodfellow, Bengio, and Courville}]{goodfellow2016deep}
Goodfellow, I., Bengio, Y., and Courville, A., \emph{Deep Learning}, MIT press, 2016.

\bibitem[{Chollet(2021)}]{chollet2021deep}
Chollet, F., \emph{Deep Learning with Python}, Simon and Schuster, 2021.

\bibitem[{Virtanen et~al.(2020)Virtanen, Gommers, Oliphant, Haberland, Reddy, Cournapeau, Burovski, Peterson, Weckesser, Bright, {van der Walt}, Brett, Wilson, Millman, Mayorov, Nelson, Jones, Kern, Larson, Carey, Polat, Feng, Moore, {VanderPlas}, Laxalde, Perktold, Cimrman, Henriksen, Quintero, Harris, Archibald, Ribeiro, Pedregosa, {van Mulbregt}, and {SciPy 1.0 Contributors}}]{2020SciPy-NMeth}
Virtanen, P., Gommers, R., Oliphant, T.~E., Haberland, M., Reddy, T., Cournapeau, D., Burovski, E., Peterson, P., Weckesser, W., Bright, J., {van der Walt}, S.~J., Brett, M., Wilson, J., Millman, K.~J., Mayorov, N., Nelson, A. R.~J., Jones, E., Kern, R., Larson, E., Carey, C.~J., Polat, {\.I}., Feng, Y., Moore, E.~W., {VanderPlas}, J., Laxalde, D., Perktold, J., Cimrman, R., Henriksen, I., Quintero, E.~A., Harris, C.~R., Archibald, A.~M., Ribeiro, A.~H., Pedregosa, F., {van Mulbregt}, P., and {SciPy 1.0 Contributors}, \enquote{{{SciPy} 1.0: Fundamental Algorithms for Scientific Computing in Python},} \emph{Nature Methods}, Vol.~17, 2020, pp. 261--272.
\newblock \doi{10.1038/s41592-019-0686-2}.

\bibitem[{Waskom(2021)}]{Waskom2021}
Waskom, M.~L., \enquote{seaborn: statistical data visualization,} \emph{Journal of Open Source Software}, Vol.~6, No.~60, 2021, p. 3021.
\newblock \doi{10.21105/joss.03021}, \urlprefix\url{https://doi.org/10.21105/joss.03021}.

\bibitem[{Feurer and Hutter(2019)}]{feurer2019hyperparameter}
Feurer, M., and Hutter, F., \emph{Hyperparameter Optimization}, Springer International Publishing, Cham, 2019, pp. 3--33.
\newblock \doi{10.1007/978-3-030-05318-5_1}, \urlprefix\url{https://doi.org/10.1007/978-3-030-05318-5_1}.

\end{thebibliography}

\end{document}